\documentclass[10pt,twocolumn,letterpaper]{article}
\usepackage{algorithm}
\usepackage{algorithmic}

\usepackage[pagenumbers]{cvpr} 

\usepackage{multirow} 
\definecolor{cvprblue}{rgb}{0.21,0.49,0.74}
\usepackage[pagebackref,breaklinks,colorlinks,allcolors=cvprblue]{hyperref}

\def\confName{CVPR}
\def\confYear{2026}
\usepackage{bm}

\title{NG-GS: NeRF-Guided 3D Gaussian Splatting Segmentation}

\author{
Yi He$^{1}$, Tao Wang$^{1}$\thanks{Corresponding author.}, Yi Jin$^{1}$, Congyan Lang$^{1}$, Yidong Li$^{1}$, Haibin Ling$^{2}$\\
$^{1}$ Key Laboratory of Big Data and Artificial Intelligence in Transportation, Ministry of Education;\\
School of Computer and Information Technology, Beijing Jiaotong University, Beijing 100044, China\\
$^{2}$ Department of Artificial Intelligence, Westlake University, Hangzhou 310030, China\\
{\small\ttfamily \{24110118, twang, yjin, cylang, ydli\}@bjtu.edu.cn, linghaibin@westlake.edu.cn}
}

\begin{document}
\maketitle
\begin{abstract}
Recent advances in 3D Gaussian Splatting (3DGS) have enabled highly efficient and photorealistic novel view synthesis. However, segmenting objects accurately in 3DGS remains challenging due to the discrete nature of Gaussian representations, which often leads to aliasing and artifacts at object boundaries. In this paper, we introduce NG-GS, a novel framework for high-quality object segmentation in 3DGS that explicitly addresses boundary discretization. Our approach begins by automatically identifying ambiguous Gaussians at object boundaries using mask variance analysis. We then apply radial basis function (RBF) interpolation to construct a spatially continuous feature field, enhanced by multi-resolution hash encoding for efficient multi-scale representation. A joint optimization strategy aligns 3DGS with a lightweight NeRF module through alignment and spatial continuity losses, ensuring smooth and consistent segmentation boundaries. Extensive experiments on NVOS, LERF-OVS, and ScanNet benchmarks demonstrate that our method achieves state-of-the-art performance, with significant gains in boundary mIoU. Code is available at \href{https://github.com/BJTU-KD3D/NG-GS}{https://github.com/BJTU-KD3D/NG-GS}.
\end{abstract}    
\section{Introduction}
\label{sec:intro}
Advancements in 3D scene~\cite{sitzmann2020sal} representation techniques have dramatically enhanced the quality of 3D view synthesis, making it possible to achieve exceptional results without the need for specialized equipment or excessive computational resources. To capitalize on these breakthroughs, it is essential to develop dedicated tools for scene understanding and manipulation, which can effectively handle these representations~\cite{zhou2025splatmesh}. 3D segmentation is a critical component for achieving accurate scene perception and interaction. Obtaining precise segmentation results is vital in neural representations, particularly those exemplified by~\textit{3D Gaussian splatting} (3DGS)~\cite{kerbl20233d}. 


\begin{figure}[t]
    \centering
    \begin{subfigure}[b]{0.115\textwidth}
        \includegraphics[width=\textwidth]{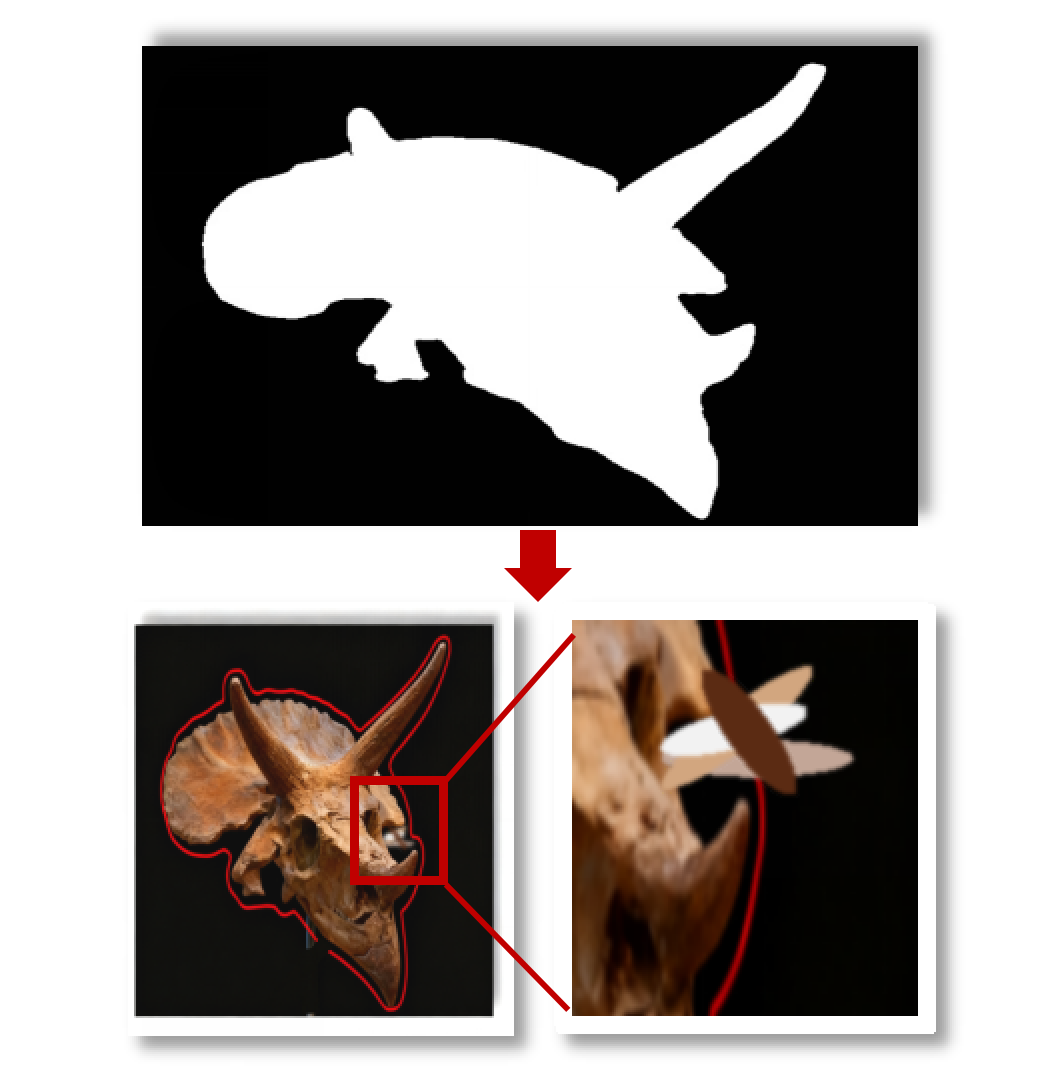}
        \caption{Mask}
        \label{fig:boundary_smoothness}
    \end{subfigure}
    \hfill
    \begin{subfigure}[b]{0.115\textwidth}
        \includegraphics[width=\textwidth]{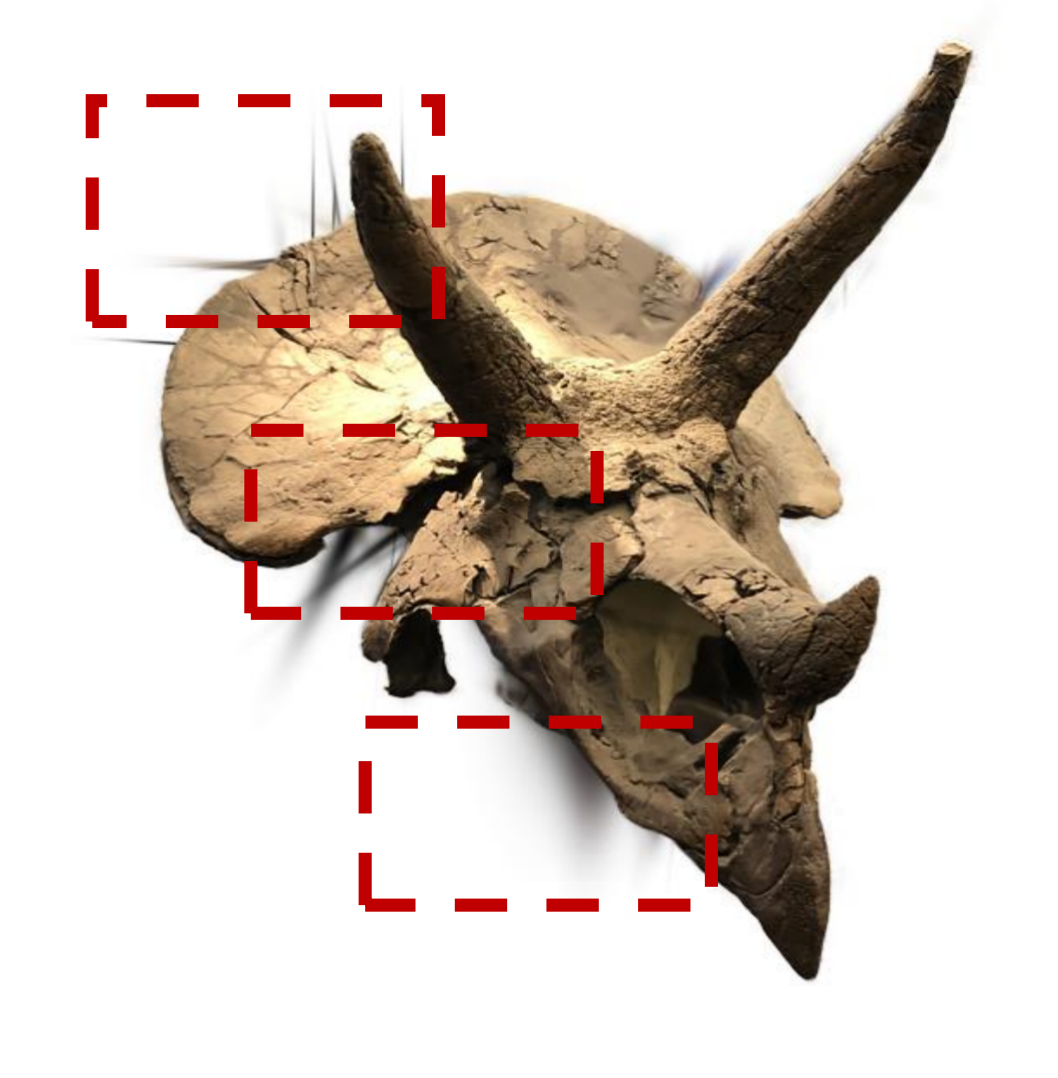}
        \caption{Mutated}
        \label{fig:novs_forns_center}
    \end{subfigure}
    \hfill
    \begin{subfigure}[b]{0.115\textwidth}
        \includegraphics[width=\textwidth]{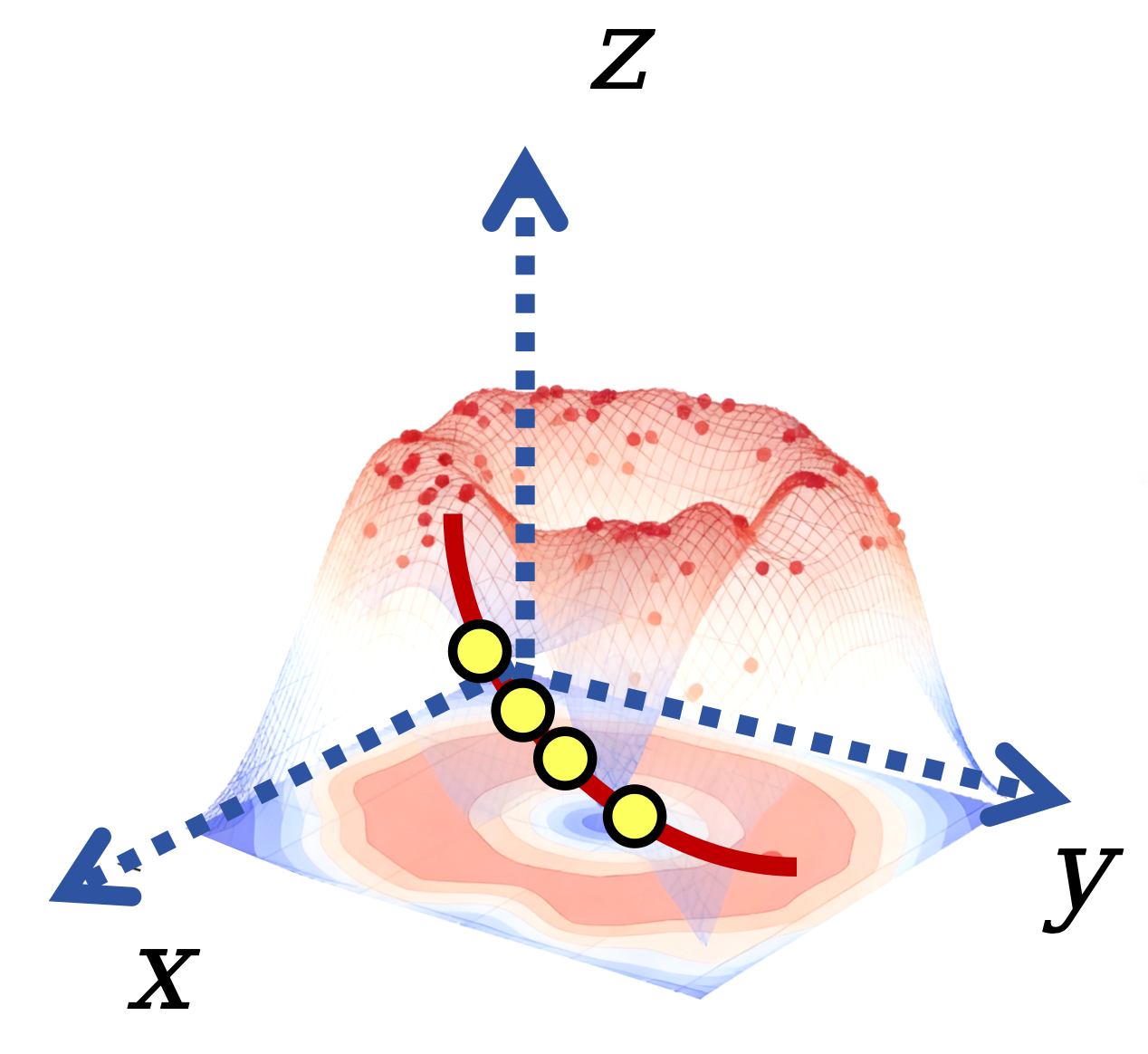}
        \caption{Continuation}
        \label{fig:mutated_gaussian}
    \end{subfigure}
    \hfill
    \begin{subfigure}[b]{0.115\textwidth}
        \includegraphics[width=\textwidth]{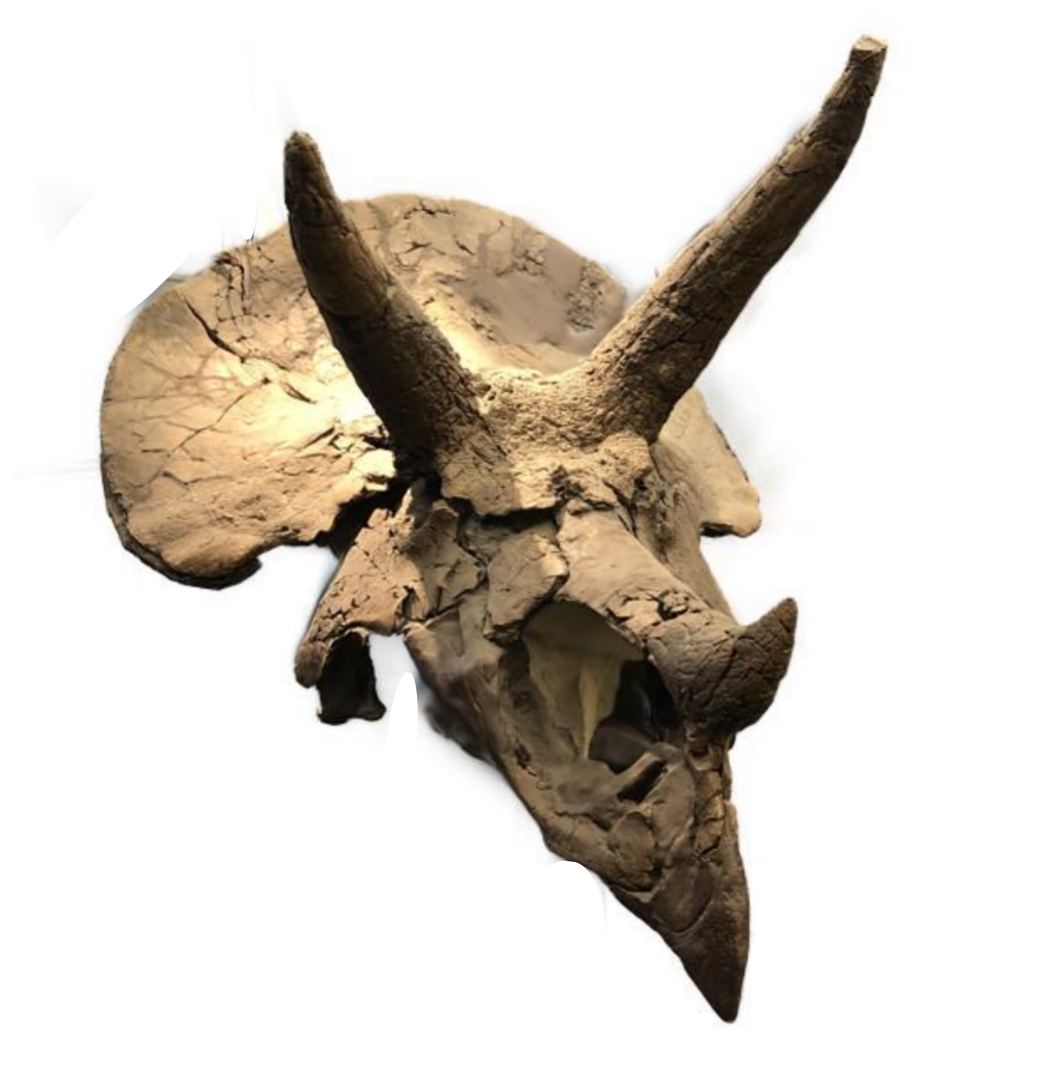}
        \caption{Our method}
        \label{fig:our_method}
    \end{subfigure}
    
    \caption{Illustrate the mutated Gaussian at the boundaries by using the mask of the object. Our method leverages the continuous representation capacity of NeRF to refine boundary Gaussians, effectively mitigating discontinuities and achieving superior segmentation performance.}
    \label{fig:boundary_comparison}
\vspace{-0.5cm}
\end{figure}

Recent 3DGS segmentation methods can be roughly categorized into feature-based, feedforward-based, and mask-based~\cite{he2025survey}. Feature-based methods~\cite{huang2023feature3dgs,peng2024vlgaussian} enhance the representation capability of edge features by extracting semantic knowledge from 2D base models (e.g. CLIP~\cite{radford2021cliptech}) into 3D scene representation. Feedforward methods~\cite{yang2024lsm, wu2024slgaussian, zhou2024drsplat} use end-to-end architectures for fast and universal semantic field construction. Recent mask-based methods~\cite{kim2024grouping, chen2024omniseg, wang2024saga} upgrade 2D masks from basic segmentation models (\textit{e.g.}, SAM~\cite{kirillov2023sam} and SAM2~\cite{kirillov2024sam2}) to 3DGS masks.
Despite these progresses, the original scene reconstruction often ignores the discreteness of Gaussian elements, especially at object boundaries. This negligence leads to the abrupt changes on the boundaries during scene segmentation, resulting in inaccurate segmentation characterized by edge artifacts, as shown in Figure~\ref{fig:boundary_comparison}. Some existing methods~\cite{li2024cobgs, tang2024sagd} directly remove the mutated boundary Gaussian distribution. However, simply removing Gaussian elements at the boundaries might interfere with visual quality.

To overcome these challenges, we propose a novel \textit{\textbf{N}eRF-\textbf{G}uided 3D\textbf{GS}} (NG-GS) segmentation framework, aiming to achieve model continuity at object boundaries. Our key insight is to utilize the continuous modeling capability of the \textit{neural radiation field} (NeRF) to adjust the coordinates and attributes of 3DGS, in order to generate spatially continuous boundaries.
Specifically, we first use the statistical variance of mask signals generated from 2D masks to automatically identify fuzzy Gaussian distributions in boundary regions. Subsequently, we use \textit{radial basis function} (RBF) interpolation to generate spatially continuous feature fields, and adopt \textit{multi-resolution hash encoding} (MRHE) to efficiently produce multi-scale representation. These features are then fed into a lightweight NeRF module as supplementary signals to enhance the spatial continuity. Finally, a joint optimization strategy is employed to coordinate 3DGS with NeRF using alignment and spatial continuity loss. Our pipeline ensures smooth transitions and consistency of segmented boundaries, significantly improving segmentation accuracy. 

For evaluation, we compare our method with state-of-the-art baseline methods on three common benchmarks, NVOS~\cite{ren2022neural}, LERF-OVS~\cite{kerr2023lerf} and ScanNet~\cite{dai2017scannet}. Experimental results reveal that our method consistently outperforms all compared baselines across all metrics on three benchmarks. The significant improvement in boundary mIoU (5.6\% on NVOS, 4.4\% on LERF-OVS, and 6.8\% on ScanNet) highlights the enhanced ability of our method in dealing with complex boundary regions.

With the proposed NG-GS framework, we make the following main contributions:
\begin{itemize}
    \item we develop a continuous feature field construction module that combines RBF interpolation with MRHE to generate spatially smooth and multi-scale feature representations, effectively addressing the discretization issue of Gaussians at boundaries; 
    \item we propose a joint optimization framework that integrates 3DGS with NeRF, leveraging alignment and spatial continuity losses to harmonize the rendering outputs of both models, thereby ensuring segmentation consistency across views; and
    \item we conduct extensive experiments showing that NG-GS achieves state-of-the-art performance, with notable gains in boundary accuracy. 
\end{itemize}

\section{Related Works}
\label{sec:formatting}
\subsection{3DGS Reconstruction}
3DGS~\cite{kerbl20233d} has recently emerged as a powerful technique for real-time neural rendering, achieving high-fidelity and photorealistic synthesis with excellent efficiency. Unlike implicit field-based methods such as NeRF~\cite{mildenhall2020nerf}, which rely on volume rendering using coordinate-based neural networks, 3DGS explicitly represents a scene as a collection of anisotropic 3D Gaussians and renders it through differentiable rasterization. While early research on 3DGS~\cite{qian20243dgs,ag20253dgs} primarily focused on novel view synthesis, recent work has expanded its application to a growing number of downstream tasks, such as simultaneous localization and mapping (SLAM)~\cite{chen2023gaussian}, human avatars~\cite{yang2023human}, segmentation~\cite{tang2023langsplat}, scene editing~\cite{liu2024editing} and content generation~\cite{zhou2024generation}. These applications demand richer representations that encompass not only geometry but also semantic information, spatial relationships, and multimodal cues. Compared to NeRF-based frameworks, 3DGS offers a more structured and interpretable representation, enabling efficient optimization, direct supervision, and intuitive manipulation. These characteristics make 3DGS particularly advantageous for advanced tasks beyond rendering.

\subsection{3DGS Scene Segmentation}
Current research on 3D semantic segmentation primarily follows three technical directions: feature-based distillation, feedforward inference, and mask-based enhancement. Feature-based methods extract semantic knowledge from 2D foundational models like CLIP~\cite{radford2021cliptech}, and distill it into 3D Gaussian representations. Early studies such as LangSplat~\cite{tang2023langsplat}, Feature3DGS~\cite{huang2023feature3dgs}, LangSurf~\cite{chen2024langsurf}, and VL-Gaussian~\cite{peng2024vlgaussian} have been developed to improve segmentation accuracy and spatial consistency. Other efforts, including high dimensional feature encoding (FMGS~\cite{zhang2024fmgs}) and decoupled compression (DF-3DGS~\cite{liu2024df3dgs}), aim to optimize storage and computational efficiency~\cite{jiang2024efficient}. Feedforward methods, such as LSM~\cite{yang2024lsm}, SLGaussian~\cite{wu2024slgaussian} and Dr-Splat~\cite{zhou2024drsplat}, establish end-to-end pipelines that directly generate 3D semantic Gaussians from sparse inputs. Mask-based enhancement methods focus on accurately and consistently lifting 2D masks, which are generated by models like SAM~\cite{kirillov2023sam} and SAM2~\cite{kirillov2024sam2}, into 3D space. Representative approaches include consistency improvement via pre-processing (Gaussian Grouping~\cite{kim2024grouping}) or post-processing (OmniSeg3D~\cite{chen2024omniseg}), end-to-end contrastive learning (SAGA~\cite{wang2024saga}), direct 3D modeling (Click Gaussian~\cite{li2024click}), and joint optimization of semantic appearance and geometry (InstanceGaussian~\cite{zhang2024instance}). 

Falling into the mask-based category, our method leverages NeRF to address the Gaussian discretization issue and improve segmentation accuracy around the object boundary.
\section{Preliminary}
\label{sec:formatting}
\begin{figure*}[h]
\centering
\includegraphics[width=1.0\textwidth,height=0.40\textheight]{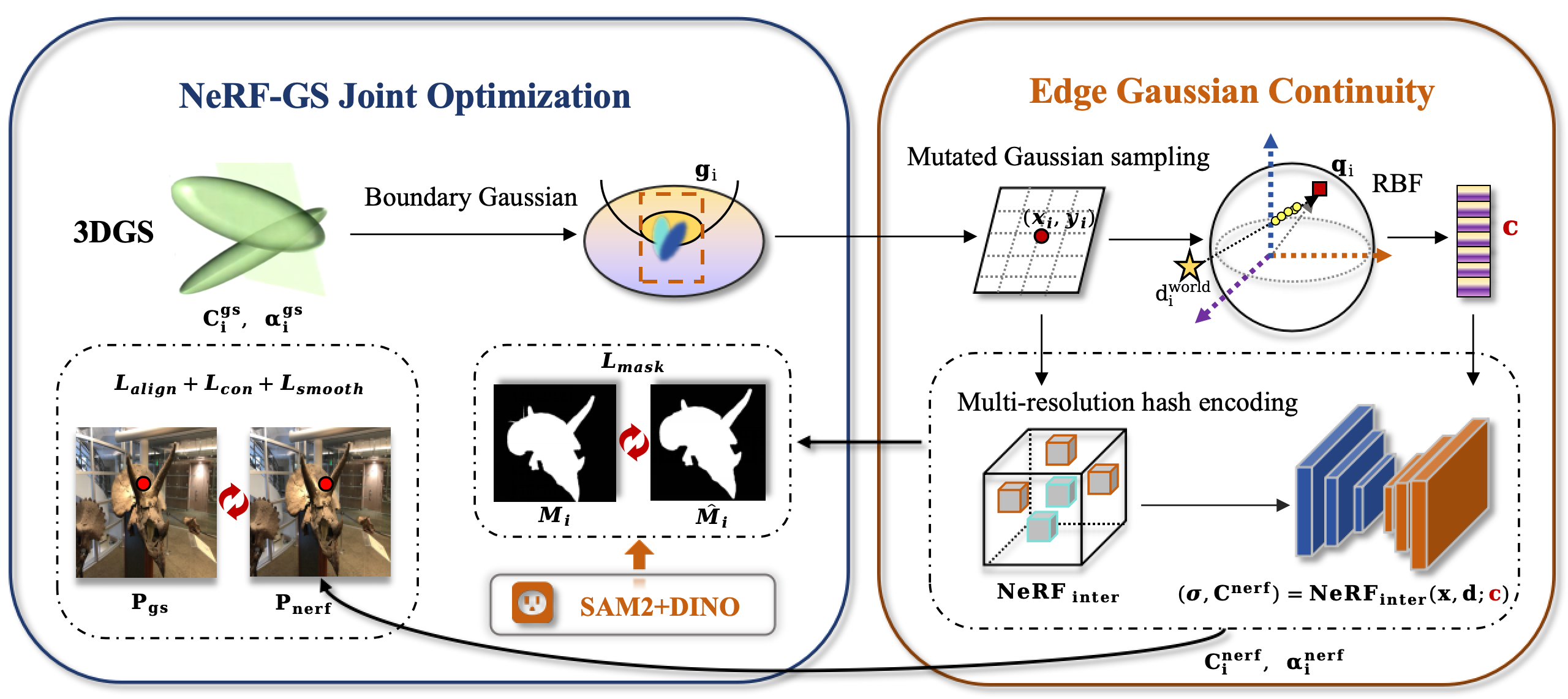} 
\caption{The overall pipeline of our NG-GS framework. It takes a trained 3DGS model as input, and identifies boundary Gaussian points with the help of a 2D segmentation model. RBF interpolation and MRHE are then used to generate continuous feature fields, and finally, boundary detail enhancement is achieved through NeRF-GS joint optimization.}
\label{fig:framework}
\end{figure*}

\subsection{NeRF}
NeRF~\cite{mildenhall2020nerf} is trained on a set of multi-view 2D images, denoted as $\mathcal{I}$. It learns a continuous function $f_{\theta}:(\mathbf{x},\mathbf{d})\to(\mathbf{c},\sigma)$, parameterized by weights $\theta$ (typically an MLP). This function maps a 3D coordinate $\mathbf{x}\in\mathbb{R}^{3}$ and viewing direction $\mathbf{d}\in\mathbb{S}^{2}$ to an RGB color $\mathbf{c}\in\mathbb{R}^{3}$ and volume density $\sigma\in\mathbb{R}_{+}$. To render a novel view, a camera ray $\mathbf{r}(t)=\mathbf{o}+t\mathbf{d}$ is cast per pixel, with $\mathbf{o}$ as the camera origin. The color $\hat{C}(\mathbf{r})$ is computed via the volume rendering integral:
\begin{equation}
\hat{C}(\mathbf{r}) = \int_{t_{n}}^{t_{f}} T(t) \, {\sigma}(\mathbf{r}(t)) \, \mathbf{c}(\mathbf{r}(t), \mathbf{d}) \, dt,
\label{eq:important}
\end{equation}
where $T(t)=\exp(-\int_{t_{n}}^{t}\sigma(\mathbf{r}(s))\,ds). \sigma(\mathbf{r}(t))$ is the transmittance, and $t_{n},t_{f}$ define the integration bounds. The term $\mathbf{c}(\mathbf{r}(t),\mathbf{d})$ denotes the local RGB color at point $\mathbf{r}(t)$ along the ray, as predicted by $f_{\theta}$.

\subsection{3DGS Segmentation}
Given a multi-view 2D image dataset with known camera poses, 3DGS~\cite{kerbl20233d} represents a scene as anisotropic 3D Gaussian primitives $\mathcal{G}=\{\mathbf{g}_{i}\}_{i=1}^{N}$. Each Gaussian also has opacity $\alpha$ and view-dependent appearance coefficients via spherical harmonics (SH). For a given camera pose, 3D-GS projects 3D Gaussians onto the 2D image plane and computes pixel color $\mathbf{c}_{\mathbf{g}_{i}}$ by blending ordered Gaussians $\mathcal{G}_{p}$ covering that pixel: 
\begin{equation}
\mathbf{C}(\mathbf{p})=\sum_{i=1}^{\left|\mathcal{G}_{\mathbf{p}}\right|}\mathbf{c}_{\mathbf{g}_{i}^{\mathbf{p}}}\alpha_{\mathbf{g}_{i}^{\mathbf{p}}}\prod_{j=1}^{i-1}\left(1-\alpha_{\mathbf{g}_{j}^{\mathbf{p}}}\right).
\end{equation}
Each Gaussian $\mathbf{g}_i$ is parameterized by mean $\bm{\mu} \in \mathbb{R}^3$ (3D position) and covariance matrix $\bm{\Sigma} \in \mathbb{R}^{3 \times 3}$ (scale and rotation, decomposed as $\bm{\Sigma} = R S S^\top R^\top$).

For 3DGS segmentation, a continuous mask label $m_i\in(0,1)$ is introduced for each Gaussian, serving as the core parameter to be learned in our framework. Similarly to the color rendering process, the mask labels of the 3D Gaussian primitives are combined through alpha compositing to yield the mask result $M$ in the 2D pixel space~\cite{li2024cobgs}.




\section{Method}
\label{sec:formatting}
The proposed \textit{NeRF-Guided 3DGS} (NG-GS) is a novel framework designed to address the challenge of object boundary segmentation in 3D Gaussian Splatting (3DGS) by mitigating the discretization artifacts that commonly occur at semantic edges.
As illustrated in Figure~\ref{fig:framework}, the framework operates in two core stages:

\begin{itemize}
\item \textit{Edge Gaussian Continuity}: We first identify ambiguous Gaussians located at object boundaries through mask variance analysis. These boundary Gaussians are then used to construct a spatially continuous feature field via RBF interpolation. To efficiently encode multi-scale spatial information, we incorporate \textit{multi-resolution hash encoding} (MRHE), which enhances the representation capacity while maintaining computational efficiency.
\item \textit{NeRF-GS Joint Optimization}: The interpolated and hash-encoded features are fed into a lightweight NeRF module, which acts as a continuous refinement network. A joint optimization strategy is employed, where alignment loss and spatial continuity loss are used to harmonize the outputs of 3DGS and NeRF. This ensures that the final segmentation maintains both high-frequency details and smooth transitions across views.
\end{itemize}

\subsection{Edge Gaussian Continuity}
This module takes the trained 3DGS model $\mathcal{G}=\{\mathbf{g}_{i}\}_{i=1}^{N_{\text{g}}}$ as input and produces spatially continuous features for the future NeRF process.

\textbf{Mutated Gaussian sampling.} Boundary Gaussian points refer to the scene boundary regions identified by analyzing the variance of the mask signals generated from Gaussian points. Specifically, for each Gaussian point $\mathbf{g}_i$ we produce a set of mask signals $\mathcal{M}_i =\left\{m_{i 1}, m_{i 2}, \ldots, m_{i N_{\text{g}}}\right\}$ through a multi-view consistency process~\cite{li2024cobgs}, where $m_{i j} \in[0,1]$ represents the foreground probability derived from segmentation models (e.g., SAM) applied to multi-view images. After computing the variance of the mask signal values $\sigma_{i}$ for each Gaussian point, the points with larger variances are considered as boundary points located in the transition region between foreground and background. Formally, we construct a boundary Gaussian set:
\begin{equation}
\mathcal{B} = \left\{ \mathbf{g}_i \mid \sigma_i^2 > \tau,\ i = 1, 2, \ldots, N_{\text{g}}\right\},    
\label{eq:boundary}
\end{equation}
 where $N_{\text{g}}$ is the total number of boundary Gaussian points.


After obtaining boundary Gaussian points, we compute their distribution range $[B_{\min},B_{\max}]$ on the image plane. We first project each boundary point $\mathbf{g}_i$ from the world coordinates $\mathbf{x}_i$ to the image coordinates $(x_i, y_i)$ using the intrinsic and extrinsic parameters for a given camera view, and then form a conservative bounding box through symmetric expansion to prevent the omission of edge points due to discrete sampling. 
\begin{equation}
\begin{aligned}
B_{\min} &= \left[ \min \{x_i \mid \mathbf{g}_i \in \mathcal{B}\} - \delta, \min \{y_i \mid \mathbf{g}_i \in \mathcal{B}\} - \delta \right], \\
B_{\max} &= \left[ \max \{x_i \mid  \mathbf{g}_i\in \mathcal{B}\} + \delta, \max \{y_i \mid \mathbf{g}_i \in \mathcal{B} + \delta \right],
\end{aligned}
\end{equation}
where $\delta $ is the extended offset, \(B_{\min}\) and \(B_{\max}\) represent the minimum and maximum boundary coordinates of the expanded bounding box, respectively. This design ensures that subsequent dense sampling can fully cover the target area, providing a spatial basis for high-quality interpolation calculations.

After that, we sample some foreground pixels from the bounding box. A uniform grid is adopted to ensure the consistency and integrity of sampling, avoiding the potential bias caused by random sampling:
\begin{equation}
\Delta x = \frac{(x_{\max} - x_{\min})} {N_{\mathrm{col}}}, \quad \Delta y = \frac{(y_{\max} - y_{\min})} {N_{\mathrm{row}}}, 
\end{equation}
where $N_{\mathrm{col}}$ and $N_{\mathrm{row}}$ are the rows and columns of the grid, respectively, and $x_{\min}$, $x_{\max}$, $y_{\min}$, $y_{\max}$ are the minimum and maximum of the x-axis and y-axis of the bounding box, respectively. The points in the sampled grid are constructed as $\mathcal{P}_{\mathrm{grid}}=\{(x_{\min}+k\Delta x,y_{\min} + l\Delta y)|0\leq k \leq N_{\mathrm{col}},0\leq l \leq N_{\mathrm{col}}\}$.

For the $i$-th sampled grid point $(x_i, y_i) \in  \mathcal{P}_{\mathrm{grid}}$, we generate a camera system direction vector using approximate internal parameters:
\begin{equation}
\begin{aligned}
\mathbf{d}_{i}^{\text{cam}} &= \operatorname{normalize}\left( \left[ \frac{x_i - c_{x}}{f_{x}}, \frac{y_i - c_{y}}{f_{y}}, 1 \right] \right), \\
\mathbf{d}_{i}^{\text{world}} &= \mathbf{R}^{T} \cdot \mathbf{d}_{i}^{\text{cam}},
\end{aligned}
\end{equation}
where $(f_x, f_y)$ are the camera focal length, $(c_x, c_y)$ denote the center coordinates of the image, and $\mathbf{R}$ is the rotation matrix that transforms the ray direction in camera coordinate system ($\mathbf{d}_{i}^{\text{cam}}$) to world coordinate system ($\mathbf{d}_{i}^{\text{world}}$).  Then, $K$ query points are sampled along the ray direction in a depth window:
\begin{equation}
\mathbf{q}_{i,k} = \mathbf{o} + \left[ \bar{t} + \max\left(  \alpha \cdot \bar{t} , \varepsilon \right) \delta_k \right] \cdot \mathbf{d}_{i}^{\text{world}} ,~~1\leq k \leq K
\end{equation}
where $\mathbf{o}$ denotes the center of the camera, $\bar{t}$ is the average depth from all boundary Gaussian points to the center of the camera, $\alpha$ (default with 0.1) determines the sampling range with $\varepsilon$ ensuring numerical stability, and the offset position $\delta_k$ is computed as $\delta_k = -1 + \frac{2(k-1)}{K-1}$.
We empirically fix $K=8$ in our experiments. By this way, we construct a query set $\mathcal{P}_{\mathrm{query}}=\{\mathbf{q}_{i,k}\}$, which consists of $N_\mathrm{row}\cdot N_\mathrm{col}\cdot K$ query points.

\vspace{1mm}\noindent\textbf{RBF interpolation.}
To solve the discontinuity problem between discrete Gaussian points, the radial basis function (RBF) is used for spatial interpolation. This method generates smooth feature transitions between boundary Gaussian points, providing continuous feature field input for NeRF.


For the $i$-th query point $\mathbf{q}_i=(x_i, y_i, z_i)$ in the query set $\mathcal{P}_{\mathrm{query}}$, we first  generate its neighbor Gaussian set $\mathcal{N}_i$ via K-NN. Subsequently, we employ RBF to compute the spatial correlation weight between the query point and its neighbor Gaussian points:
\begin{equation}
w_{i,j} = \exp\Big(-\frac{\left\|\mathbf{q}_i - \mathbf{x}_{j}\right\|^{2}}{2\sigma^{2}}\Big), \quad j \in \mathcal{N}_i,
\end{equation}
where $\mathbf{x}_j$ denotes the world coordinates of the Gaussian point $j$, and $\sigma^{2}$ serves as the width of the kernel controlling the smoothness of the interpolation.
After applying an $L_{1}$ normalization on these weights, we use them to form the interpolation feature for each query point:
\begin{equation}
    \mathbf{f}^{\mathrm{inter}}_i = \sum_{j \in\mathcal{N}_i} w_{i,j} \mathbf{f}_j,
\end{equation}
where $\mathbf{f}_j$ denotes the learnable parameters related to the Gaussian point $j$, which is initialized as its color attributes.
Through RBF interpolation, the discrete Gaussian features are fused into continuous features $\mathbf{f}^{\mathrm{inter}}$, which are then fed into the NeRF module to reinforce spatial coherence across the representation.

\vspace{1mm}\noindent\textbf{Multi-resolution hash encoding} is an efficient positional encoding technique that replaces traditional trigonometric function encoding and achieves fast feature lookup through multi-level hash tables~\cite{muller2022instantren2022neural}. We use it to replace the traditional trigonometric function encoding in NeRF to extract high-dimensional features of ray upsampling points.

The design goal of hash functions is to achieve efficient mapping from spatial coordinates to hash table indices. By multiplying coordinate components with prime numbers and XORing, we can fully utilize three-dimensional spatial information to generate unique hash values.
\begin{equation}
\operatorname{hash}(\mathbf{x}, l) = \left( \bigoplus_{i=1}^{3} x_{i} \pi_{i} \right) \bmod T_{l},
\end{equation}
where $x_{i}$ denotes the $i$-th channel of the Gaussian coordinate $\mathbf{x}$. The hash function employs a bitwise XOR operation $\oplus$ on the coordinate components $x_i$ and the corresponding prime numbers \(\pi_i\). The result is modulated by \(T_l\), which is the size of the hash table at the $l$-th level.

We capture a complete information spectrum from global structure to local details by performing hash encoding and feature search at different resolution levels. The concatenation operation combines these features of different scales into a unified representation, providing rich input information for subsequent neural networks.
\begin{equation}
\begin{split}
\mathbf{f}^{\mathrm{hash}} = \operatorname{concat} \bigl( 
    &\operatorname{lookup}(\operatorname{hash}(\mathbf{x}, 0)), \ldots,\\
    &\operatorname{lookup}(\operatorname{hash}(\mathbf{x}, L-1)) \bigr),
\end{split}
\end{equation}
where $\operatorname{lookup}$ refers to the hash lookup operation that retrieves the corresponding feature vector from the hash table based on the hash index. The final feature $\mathbf{f}^{\text{hash}}$ is a concatenation of multi-level features, capturing spatial information from coarse to fine. This feature is combined with interpolation features $\mathbf{f}^{\mathrm{inter}}$ to form the input of NeRF.

\subsection{NeRF-GS Joint Optimization}
NeRF projects a 3D scene onto a 2D image using volume rendering equations and calculates pixel colors using continuous integration. This process enhances details in the boundary area and blends with the 3DGS rendering results.
Hash encoding feature $\mathbf{f}^{\mathrm{hash}}$ serves as the main input for the NeRF network:
\begin{equation}
    \mathbf{h}^{(0)} = \operatorname{ReLU}\left(\mathbf{W}^{(0)} \mathbf{\mathbf{f}^{\mathrm{hash}}} + \mathbf{b}^{(0)}\right),
\end{equation}
where $\mathbf{W}^{(0)}$ is a weight matrix and $\mathbf{b}^{(0)}$ a bias vector.

The interpolated feature $\mathbf{f}^{\mathrm{inter}}$ is input as a condition vector \(\mathbf{c}\) of the NeRF network. Different from the original MLP in NeRF, our network incorporates interpolation features to modulate the activation function of each layer. The condition vector \(\mathbf{c}\) is linearly transformed into the modulation parameters \(\gamma^{(l)}\) and $\beta^{(l)}$, which modulate the activation vectors of all layers:
\begin{equation}
    \gamma^{(l)} = \mathbf{W}_{\gamma}^{(l)} \mathbf{c} + \mathbf{b}_{\gamma}^{(l)}, \quad \beta^{(l)} = \mathbf{W}_{\beta}^{(l)} \mathbf{c} + \mathbf{b}_{\beta}^{(l)},
\end{equation}
where $\mathbf{W}_{\gamma}^{(l)}$ and $\mathbf{W}_{\beta}^{(l)}$ are learnable weight matrices and $\mathbf{b}_{\gamma}^{(l)}$ and $\mathbf{b}_{\beta}^{(l)}$ are bias terms. 
These parameters dynamically adjust the hidden layers based on external conditions.
\begin{equation}
    \mathbf{\hat{h}}^{(l)} = \mathrm{ReLU}\left( \gamma^{(l)} \odot \mathbf{h}^{(l)} + \beta^{(l)} \right),
\end{equation}
where $\odot$ is the element-wise product, $\mathbf{h}^{(l)}$ is the original activation vector of the current layer, and $\mathbf{\hat{h}}_{(l)}$ is the output activation vector after modulation of the interpolation feature, which serves as the input of the next layer.  

The final output of the modified NeRF network predicts volume density \( \sigma \) and RGB color \( \mathbf{C}^{\text{nerf}} \):
\begin{equation}
(\sigma, \mathbf{C}^{\text{nerf}}) = \mathrm{NeRF_{inter}}(\mathbf{x}, \mathbf{d}; \mathbf{c}),
\end{equation}
where $\mathbf{x}$ denotes the world coordinates of the boundary Gaussian points, and $\mathbf{d}$ is obtained through SfM.


\vspace{1mm}\noindent\textbf{Loss functions.} 
The boundary alignment loss function is a key constraint to ensure the coordinated operation of NeRF and 3DGS in the boundary region, which enforces consistency by penalizing the differences in color and transparency between the two models.
\begin{equation}
\begin{split}
\mathcal{L}_{\text{align}} 
= \frac{1}{|\mathcal{B}|} & \sum_{i \in \mathcal{B}} \Big[ \left\|\mathbf{C}^{\text{gs}}_{i}-\mathbf{C}^{\text{nerf}}_{i}\right\|_{2}^{2} +   \\
 & w_\alpha \big(\alpha^{\text{gs}}_{i} - \alpha^{\text{nerf}}_{i}\big)^{2} + w_v \operatorname{Var}\left(\mathcal{R}_{i}\right) \Big],
\end{split}
\end{equation}
where $\mathcal{B}$ represents the set of pixels in the boundary region, which is determined through semantic analysis and requires special processing; $\mathbf{C}^{\text{gs}}_{i}$ and $\mathbf{C}^{\text{nerf}}_{i}$ represent the RGB color vectors predicted by 3DGS and NeRF at the $i$-th boundary points, respectively; $\alpha^{\text{gs}}_{i}$ and $\alpha^{\text{nerf}}_{i}$ correspond to the opacity values predicted by the two models; $w_\alpha$ and $w_v$ are a hyperparameter that balances the relative importance between color loss and transparency loss. For pixel $i$, $R_i$ denotes a $7 \times 7$ pixel block around it, and $\operatorname{Var(\cdot)}$ calculates the color variance of the pixel block.

A continuity loss is explored to measure the consistency between two close boundary Gaussian points:
\begin{equation}
\mathcal{L}_{\text{cont}} = \frac{1}{|\mathcal{B}|}\sum_{i\in\mathcal{B}}(\frac{1}{|\mathcal{N}_i|}\sum_{j\in\mathcal{N}_i}\left\|\mathbf{C}^{\mathrm{gs}}_{i}-\mathbf{C}^{\mathrm{gs}}_{j}\right\|_{2}^{2}),
\end{equation}
where $\mathcal{N}_i$ represents the neighbor points of boundary point $i$, which are usually determined by a \textit{K-nearest neighbor} (K-NN) algorithm.  

The gradient smoothness loss function achieves visual smoothness by minimizing the magnitude of color gradients, thereby penalizing abrupt color variations.
\begin{equation}
\mathcal{L}_{\text{smth}} = \frac{1}{|\mathcal{B}|}\sum_{i=1}^{\mathcal{B}}\left\|\nabla\mathbf{g}_{i}\right\|_{2}^{2},
\end{equation}
where \(\nabla \mathbf{g}_i\) is calculated by a vector Jacobi product on the scalarization output of each color channel. This loss penalizes gradient mutations and enhances visual smoothness.

Similar to COB-GS, we use a mask loss to supervise the mask label training process, but add NeRF density generation weights to optimize edge Gaussian continuity learning:
\begin{equation}
\begin{aligned}
\mathcal{L}_{\text{mask}} =  \frac{1}{N_{\mathrm{px}}}  \sum_{i=1}^{N_{\mathrm{px}}} & \mathrm{sigmoid}(d_i)  \Big[  -M_{i} \log ( \hat{M}_{i}+\varepsilon) \\
&  - \left(1-M_{i}\right) \log (1-\hat{M}_{i}+\varepsilon ) \Big]
\end{aligned}
\end{equation}
where $N_{\mathrm{px}} $ is the number of pixels in the image, $d_i $ is the volumetric density of the $i$-th pixel rendered by NeRF,  $\hat{M}_{i} $ is the groundtruth image mask label, and $M_ {i}$ is the predicted value of the rendering mask.

The overall loss function during the joint optimization period is as follows:
\begin{equation}
\begin{aligned}
\mathcal{L}_{\text{total}} = \mathcal{L}_{\text{align}} &+ \lambda_m\mathcal{L}_{\text{mask}} + \lambda_c\mathcal{L}_{\text{cont}} + \lambda_s\mathcal{L}_{\text{smth}}, 
\end{aligned}
\end{equation}
where $\lambda_m$, $\lambda_c$, and $\lambda_s$ are balance factors and set as 0.5, 0.1, and 0.05, respectively.

\section{Experiments}
\label{sec:formatting}

\subsection{Implementation Details}
\textbf{Dataset.} For evaluation, we use the Neural Volumetric Object Selection (NVOS)~\cite{ren2022neural}, LERF-OVS~\cite{kerr2023lerf}, and ScanNet~\cite{dai2017scannet}. NVOS consists of eight scenes picked from the LLFF~\cite{mildenhall2019local} dataset. For each scene, NVOS provides a reference view with graffiti and a target view with annotated 2D segmentation masks. Similar to NVOS, LERF-OVS provides manual annotation for three scenes to evaluate the performance of interactive 3D segmentation. ScanNet provides posed RGB images from video scans, as well as reconstructed point clouds and 3D point-level semantic labels.

\vspace{1mm}\noindent\textbf{Baselines.} The proposed method is compared against a range of state-of-the-art baselines, which are categorized into mask-based and feedforward-based approaches. The mask-based methods include SA3D-GS~\cite{cen2025segment}, SAGA~\cite{wang2024saga}, FlashSplat~\cite{chen2024flashsplat} and COB-GS~\cite{li2024cobgs}, which focus on enhancing segmentation through geometric priors or joint optimization but often struggle with boundary discretization. The feedforward-based methods such as LSeg~\cite{liu2023lseg}, LangSplat~\cite{tang2023langsplat}, LangSurf~\cite{chen2024langsurf}, LSM~\cite{yang2024lsm}, and LanScene-X~\cite{zhang2024lanscene}, leverage end-to-end pipelines for efficient inference. 

\vspace{1mm}\noindent\textbf{Implementation details.} 
Quantitative experiments are conducted on an NVIDIA RTX 3090 GPU with PyTorch, focusing on metrics including mIoU, mAcc, and \textit{boundary mIoU} (B-mIoU). The optimizers for both 3DGS and NeRF employ Adam, with an initial learning rate of $1.6e^{-4}$. The segmentation threshold $\tau$ and the boundary box expansion offset $\delta $ are set to 0.6 and 5 pixels, respectively.

\subsection{Quantitative Results}

The quantitative results (Table~\ref{tab_NVOS}-Table~\ref{ScanNet}) show that our method outperforms all baselines across all metrics on the NVOS, LERF-OVS, and ScanNet datasets. 

Table~\ref{tab_NVOS} on NVOS shows a clear performance trajectory for mask-based segmentation, and our method represents the latest technology currently available. Similarly, Table~\ref{tab_LERF-OVS} on LERF-OVS reveals a performance hierarchy: feedforward methods (e.g., LSeg, LangSplat) exhibit limited capability, mask-based methods (e.g., SA3D-GS, COB-GS) perform better, and our method establishes a new state-of-the-art technology. Furthermore, as shown in Table~\ref{ScanNet}, NG-GS can be directly applied to multi-object segmentation since the boundary set $\mathcal{B}$ is independently computed per object and then jointly optimized.

\begin{table}[!t]
\setlength{\tabcolsep}{6.9pt} 
\centering
\caption{Performance comparison (\%) on NVOS dataset.}
\label{tab:3dgs_methods_comparison}
\begin{tabular}{lccc}
\toprule
\textbf{Method} & \textbf{B-mIoU} & \textbf{mAcc} & \textbf{mIoU} \\
\midrule
SA3D-GS~\cite{cen2025segment} & 75.6 & 98.3 & 90.7 \\
SAGA~\cite{wang2024saga} & 76.4 & 98.3 & 90.9 \\
FlashSplat~\cite{chen2024flashsplat} & 78.2 & 98.6 & 91.8 \\
COB-GS~\cite{li2024cobgs} & 79.1 & 98.6 & 92.1 \\
\hline
Ours & \textbf{84.7} & \textbf{99.2} & \textbf{92.6} \\
\bottomrule
\end{tabular}
\vspace{-0.2cm}
\label{tab_NVOS}
\end{table}

\begin{table}[!t]
\setlength{\tabcolsep}{6.5pt} 
\centering
\caption{Performance comparison (\%) on LERF-OVS dataset.}
\label{tab:performance_comparison_overall}
\begin{tabular}{lccc}
\toprule
\textbf{Method} & \textbf{B-mIoU} & \textbf{mAcc} & \textbf{mIoU} \\
\midrule
LSeg~\cite{liu2023lseg} & 30.1 & 64.1 & 39.9 \\
LangSplat~\cite{tang2023langsplat} & 10.6 & 35.7 & 15.3 \\
LangSurf~\cite{chen2024langsurf} & 12.6 & 40.2 & 19.7 \\
LSM~\cite{yang2024lsm} & 14.3 & 49.6 & 22.8 \\
LanScene-X~\cite{zhang2024lanscene} & 42.2 & 80.8 & 50.5 \\
\hline
SA3D-GS~\cite{cen2025segment} & 65.3 & 83.3 & 80.4 \\
SAGA~\cite{wang2024saga} & 66.3 & 83.3 & 80.7 \\
FlashSplat~\cite{chen2024flashsplat} & 67.2 & 83.6 & 81.3 \\
COB-GS~\cite{li2024cobgs} & 68.4 & 84.1 & 82.4 \\
\hline
Ours & \textbf{72.8} & \textbf{85.5} & \textbf{82.9} \\
\bottomrule
\end{tabular}
\vspace{-0.2cm}
\label{tab_LERF-OVS}
\end{table}

\begin{table}[!t]
\setlength{\tabcolsep}{7.6pt}
\centering
\caption{Performance comparison (\%) on  ScanNet dataset.}
\label{ScanNet}
\begin{tabular}{lccc}
\toprule
{\textbf{Method}} & {\textbf{B-mIoU}} & {\textbf{mAcc}} & {\textbf{mIoU~}}  \\
\midrule
COB-GS~\cite{li2024cobgs} & 52.8 & 80.2 & 61.6  \\
Ours & \textbf{59.6} & \textbf{84.1} & \textbf{64.3} \\
\bottomrule
\end{tabular}
\vspace{-0.5cm}
\end{table}


\begin{figure*}[!t]
\centering
\small
\begin{tabular}{@{}c@{\hspace{0.5em}}c@{\hspace{0.5em}}c@{\hspace{0.5em}}c@{\hspace{0.5em}}c@{\hspace{0.5em}}c@{}}
\rotatebox{90}{\hspace{0.6cm} fern} &
\includegraphics[width=.195\linewidth]{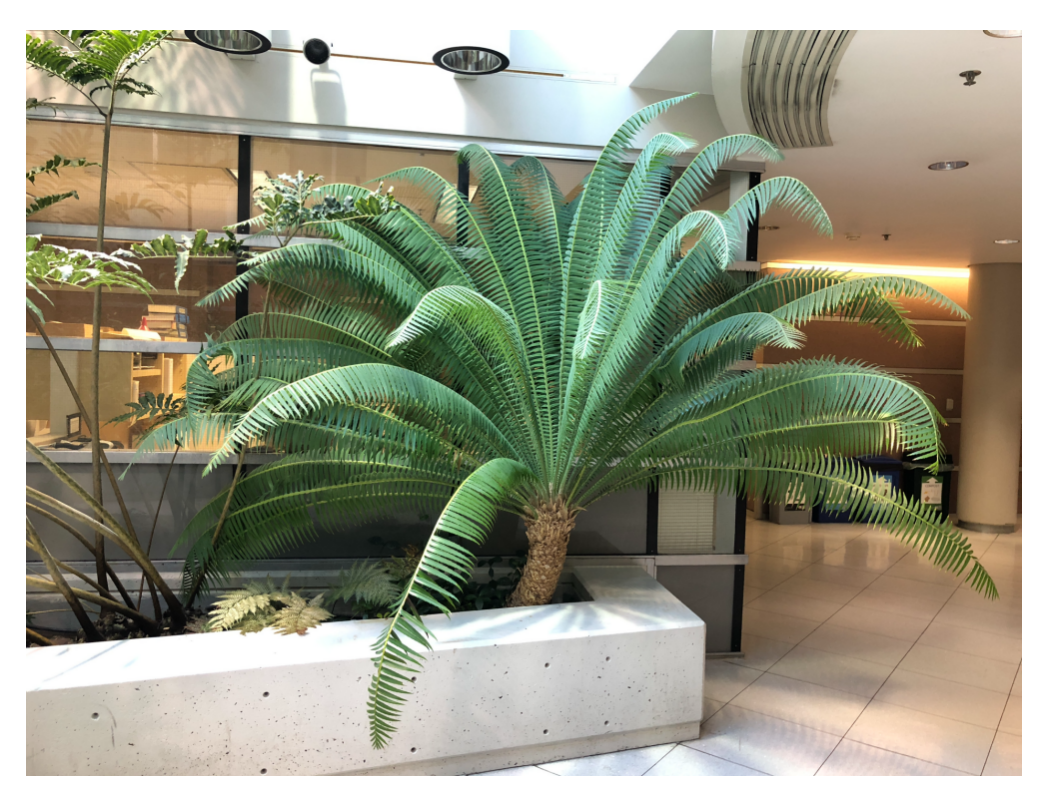} &
\includegraphics[width=.195\linewidth]{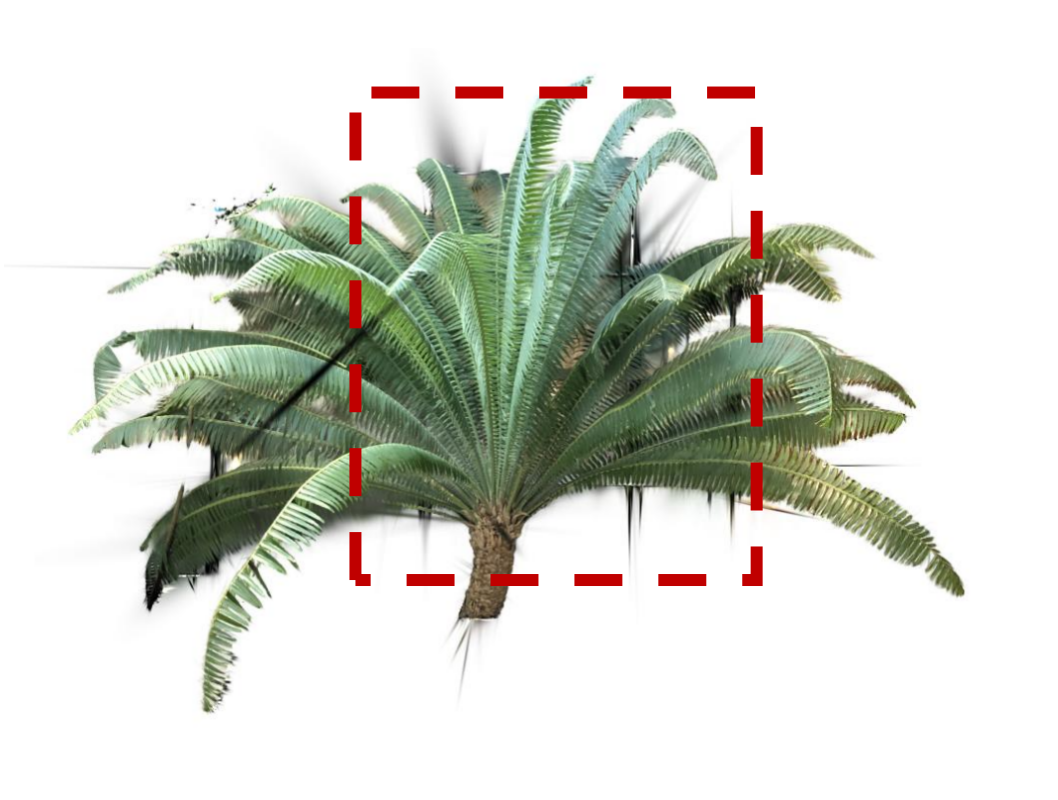} &
\includegraphics[width=.195\linewidth]{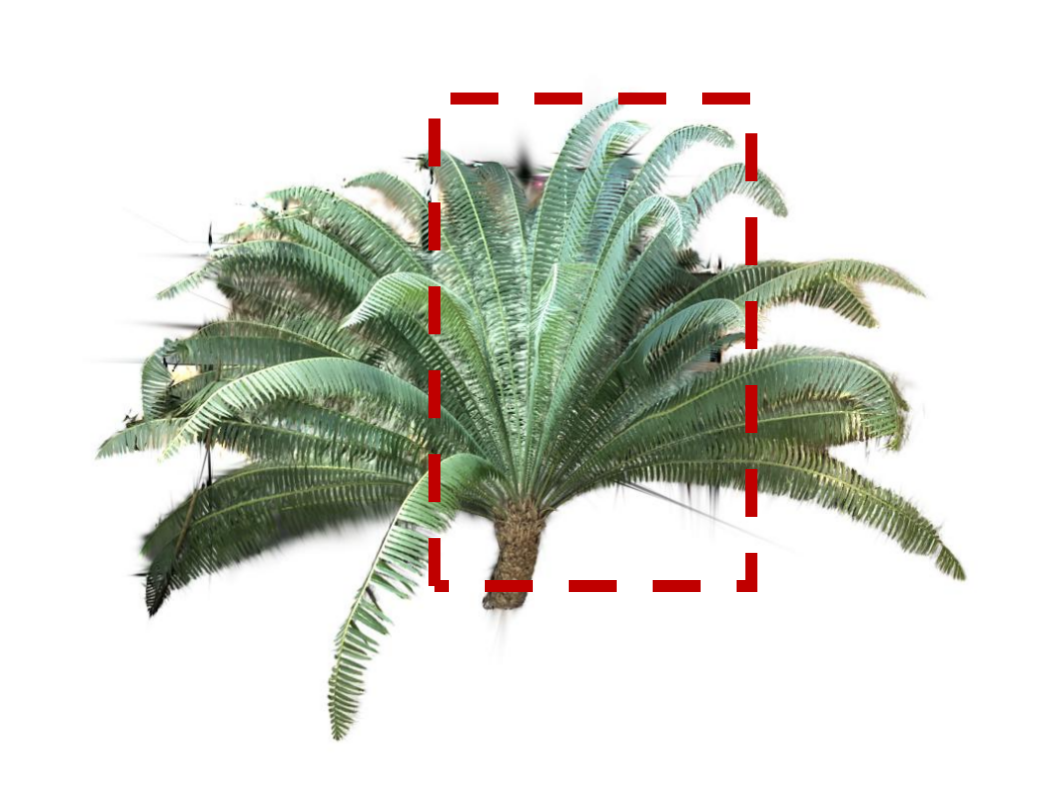} &
\includegraphics[width=.195\linewidth]{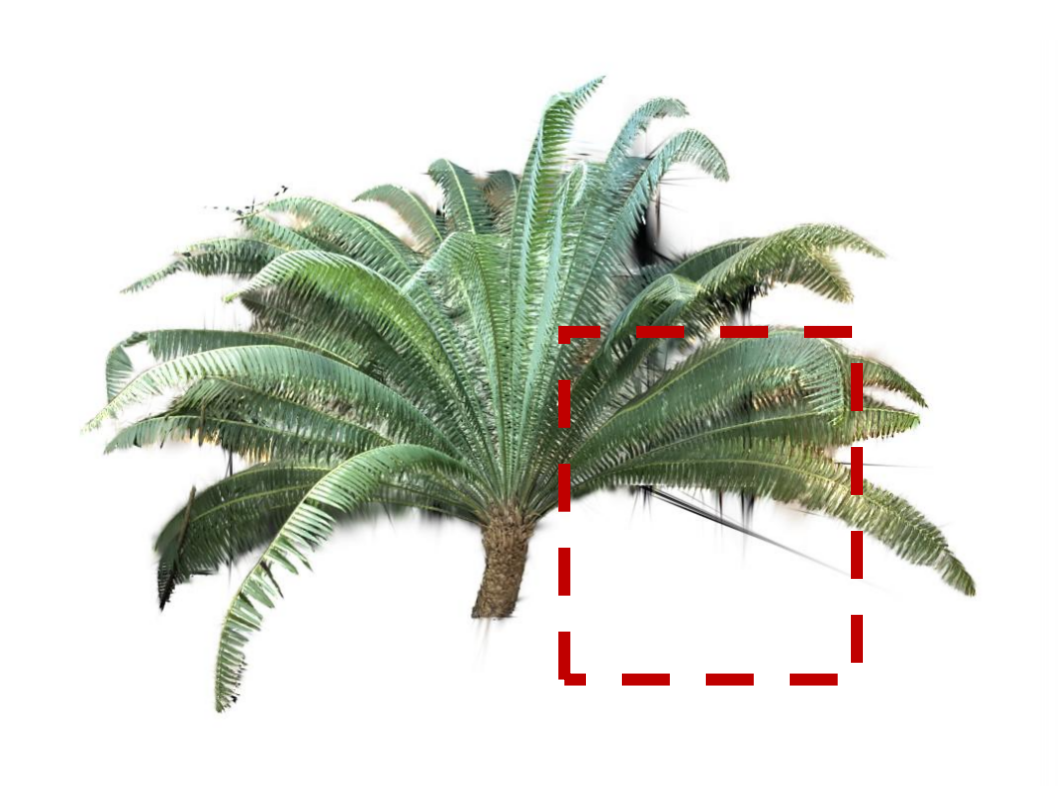} &
\includegraphics[width=.195\linewidth]{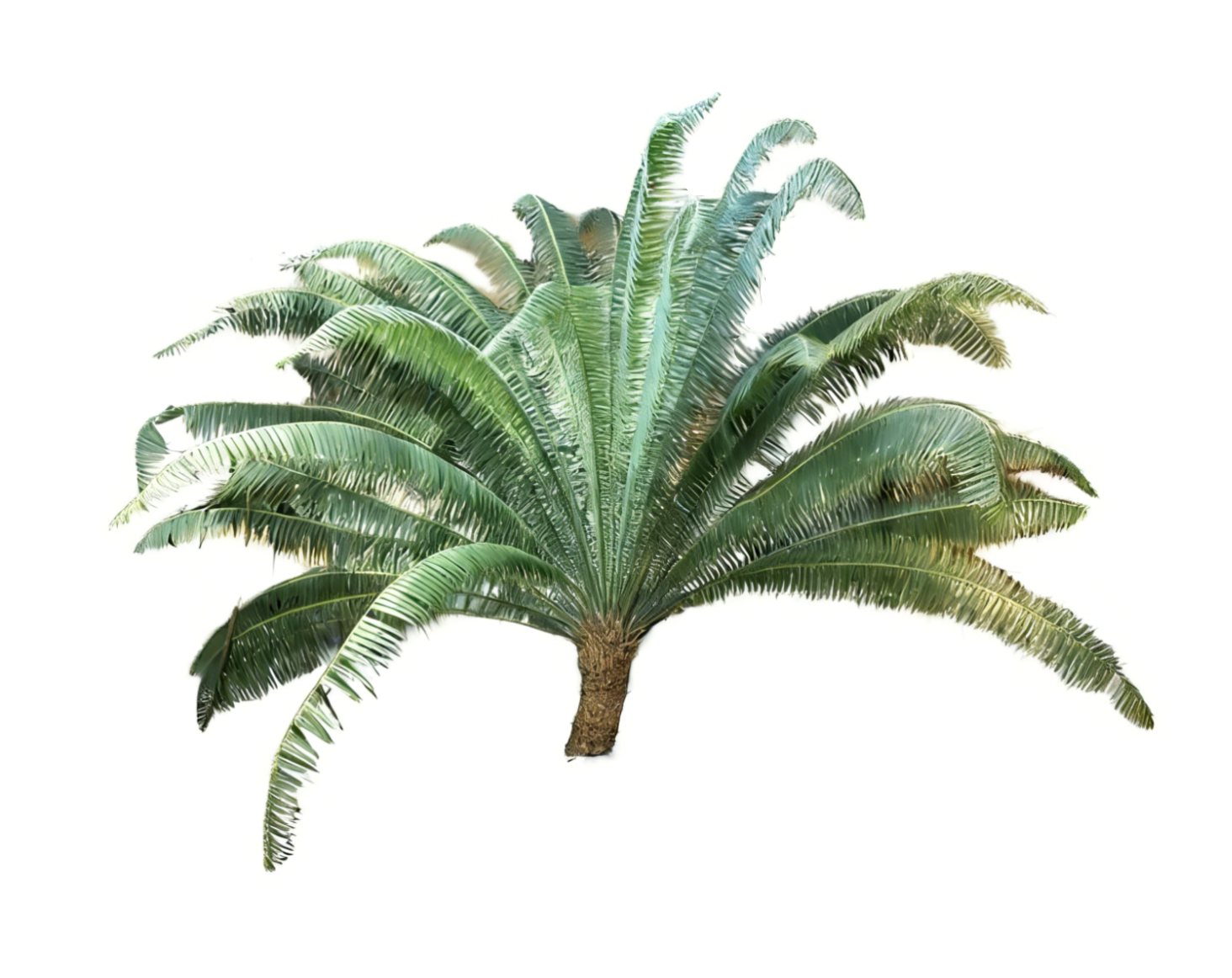} \\
\rotatebox{90}{\hspace{0.6cm} horns } &
\includegraphics[width=.195\linewidth]{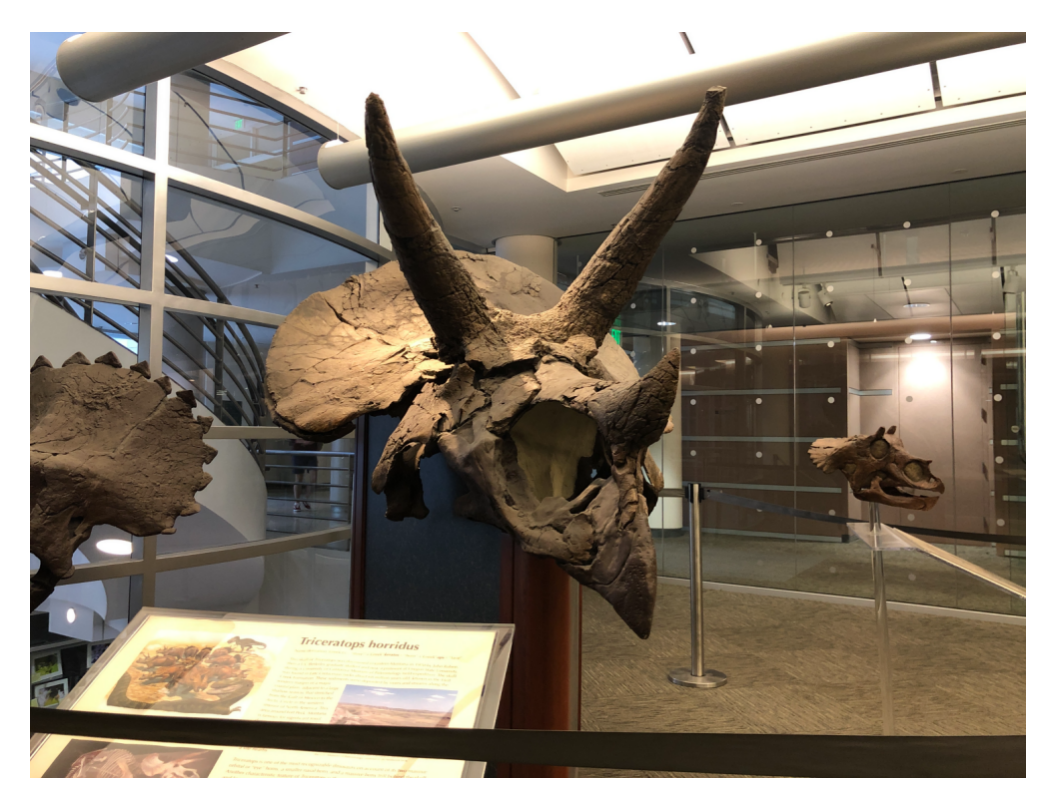} &
\includegraphics[width=.195\linewidth]{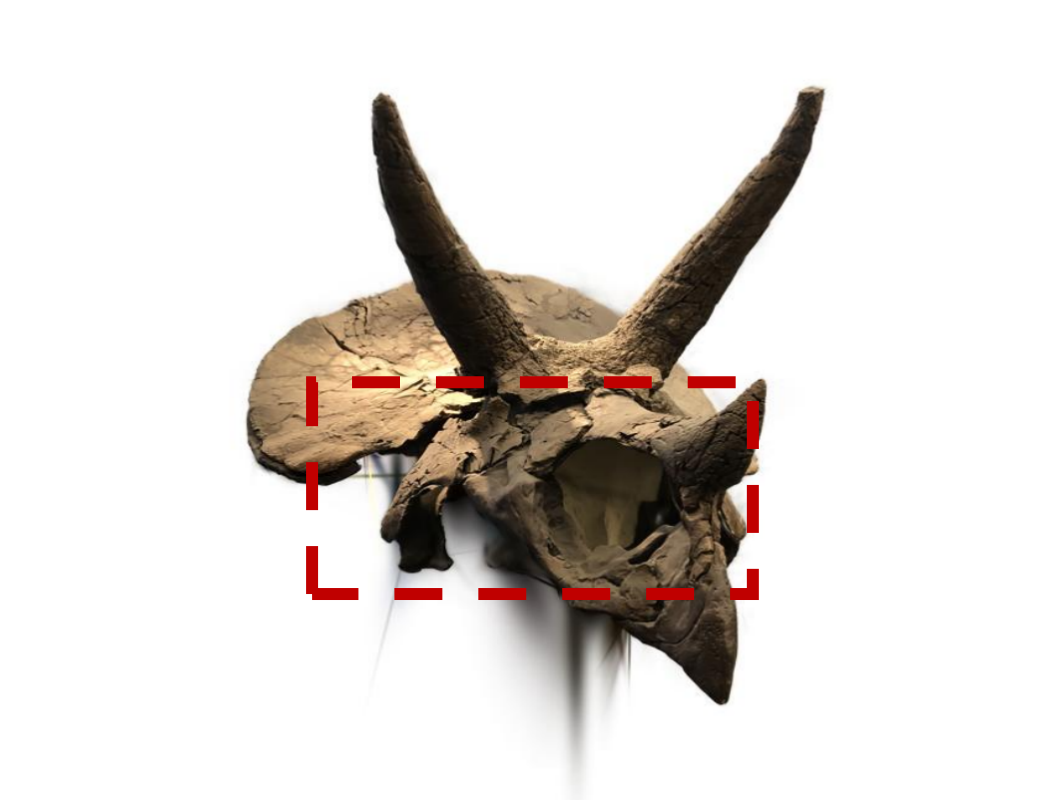} &
\includegraphics[width=.195\linewidth]{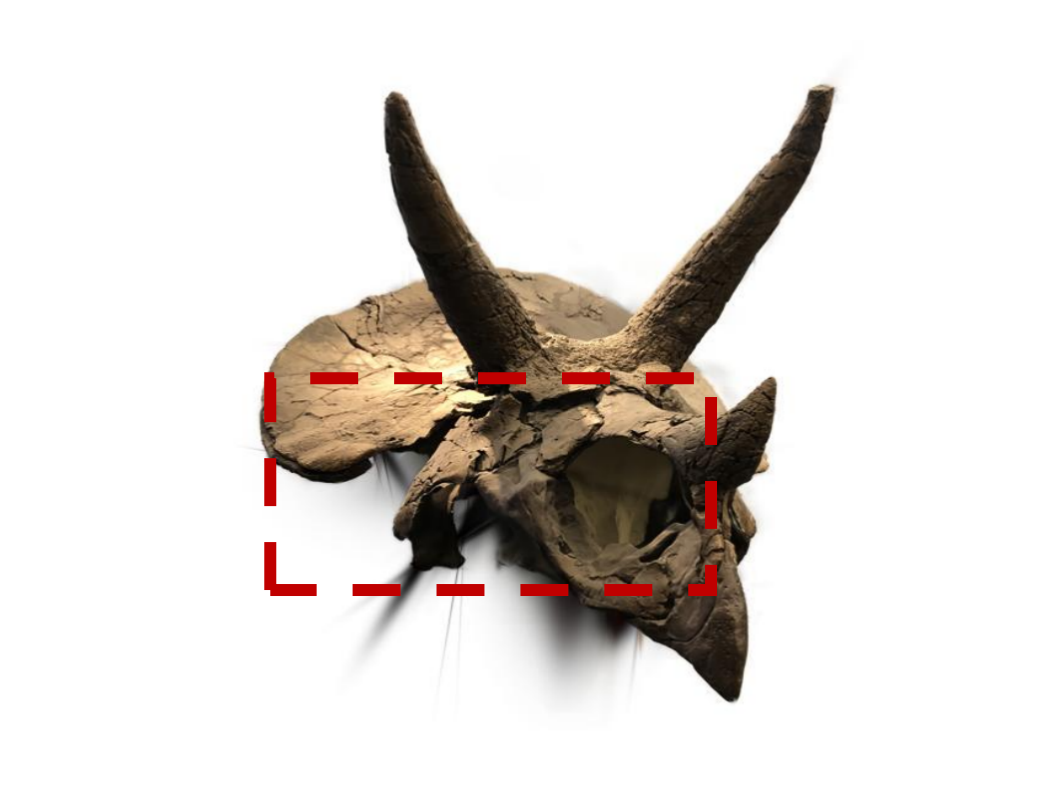} &
\includegraphics[width=.195\linewidth]{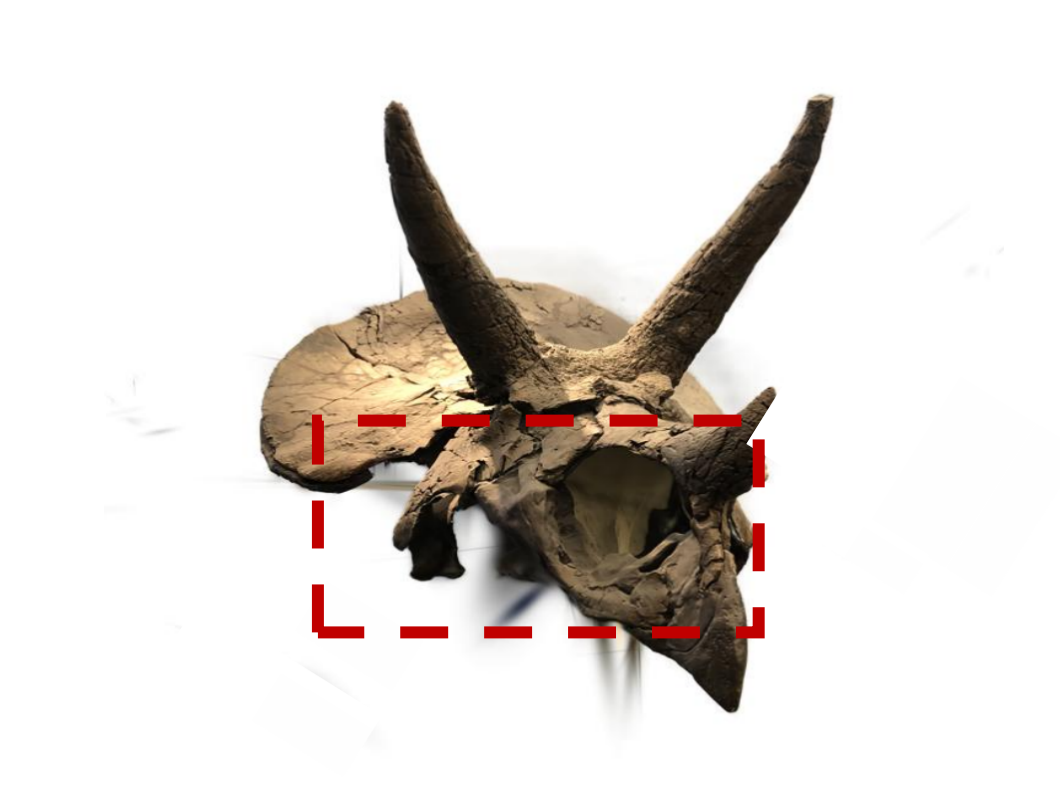} &
\includegraphics[width=.195\linewidth]{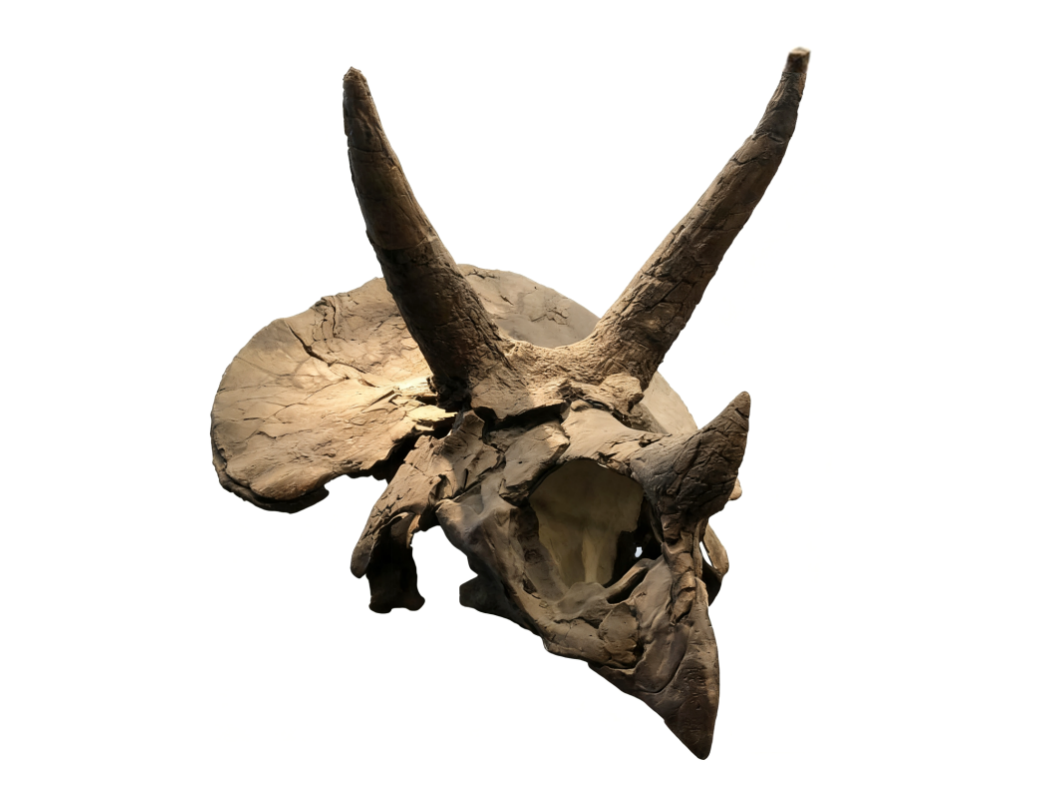} \\
\rotatebox{90}{\hspace{0.6cm} figurines} &
\includegraphics[width=.195\linewidth]{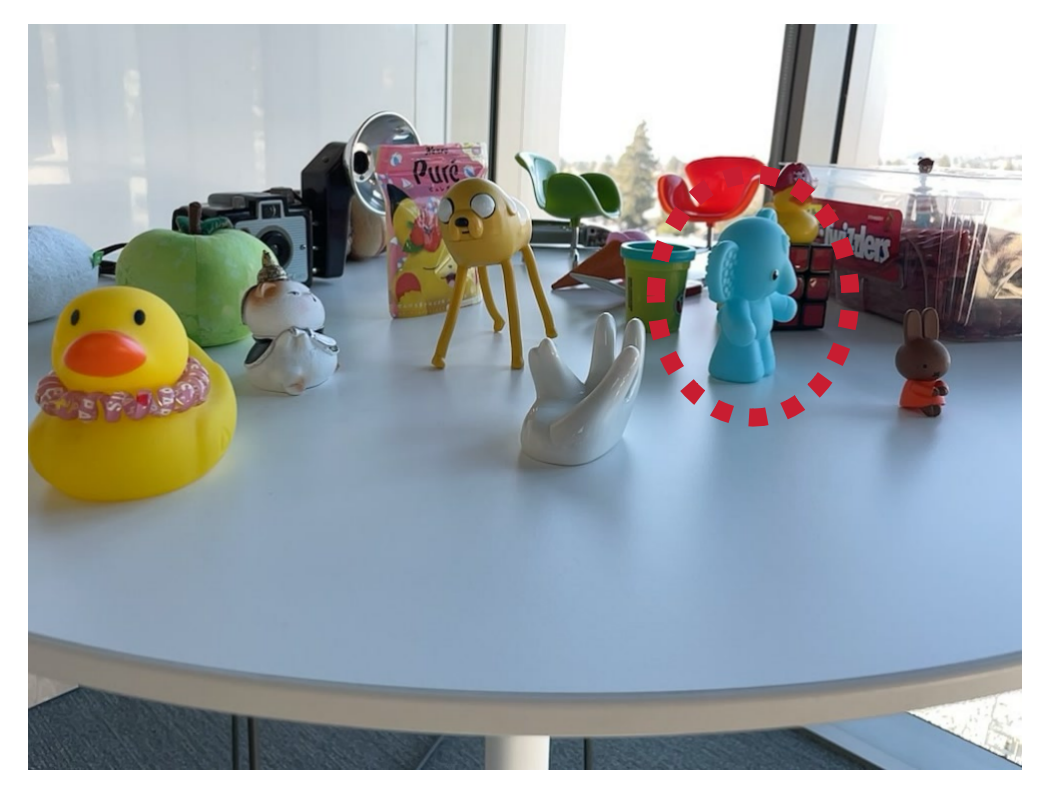} &
\includegraphics[width=.195\linewidth]{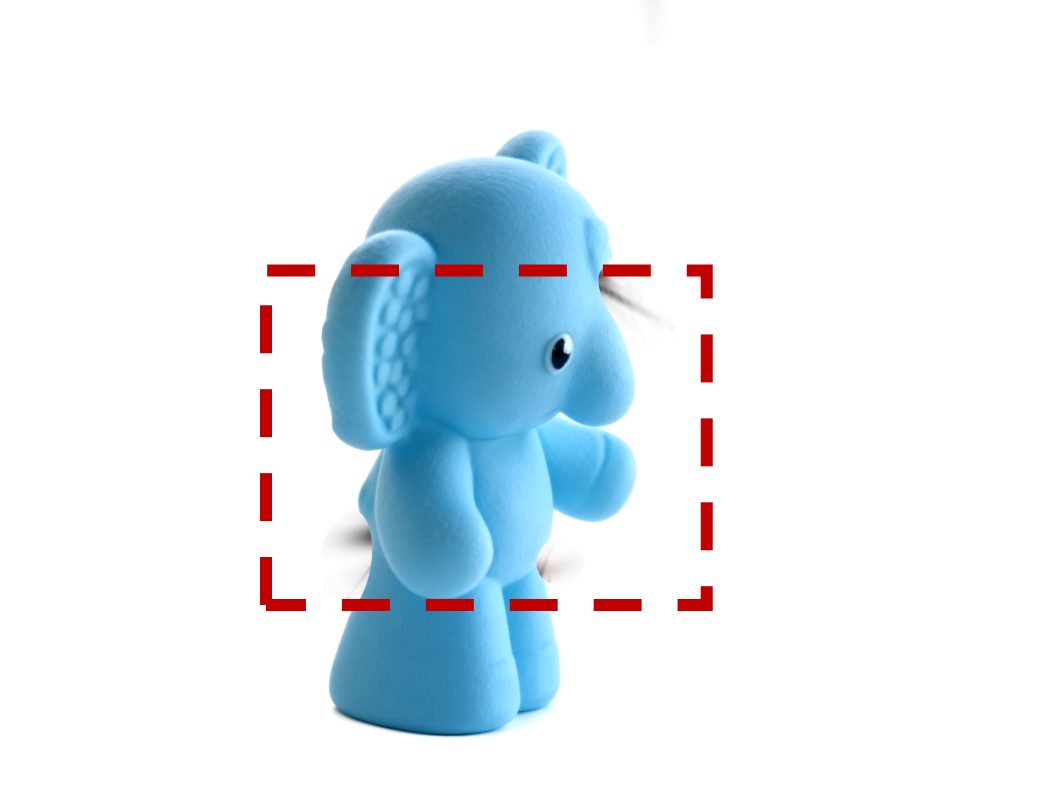} &
\includegraphics[width=.195\linewidth]{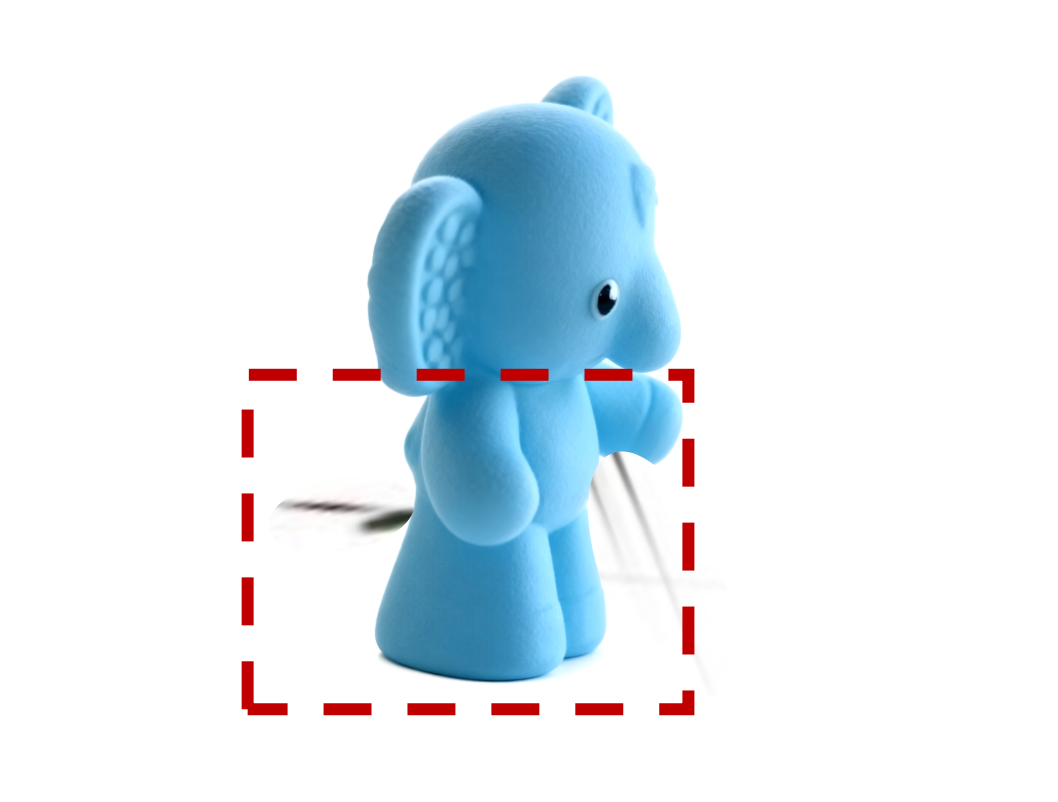} &
\includegraphics[width=.195\linewidth]{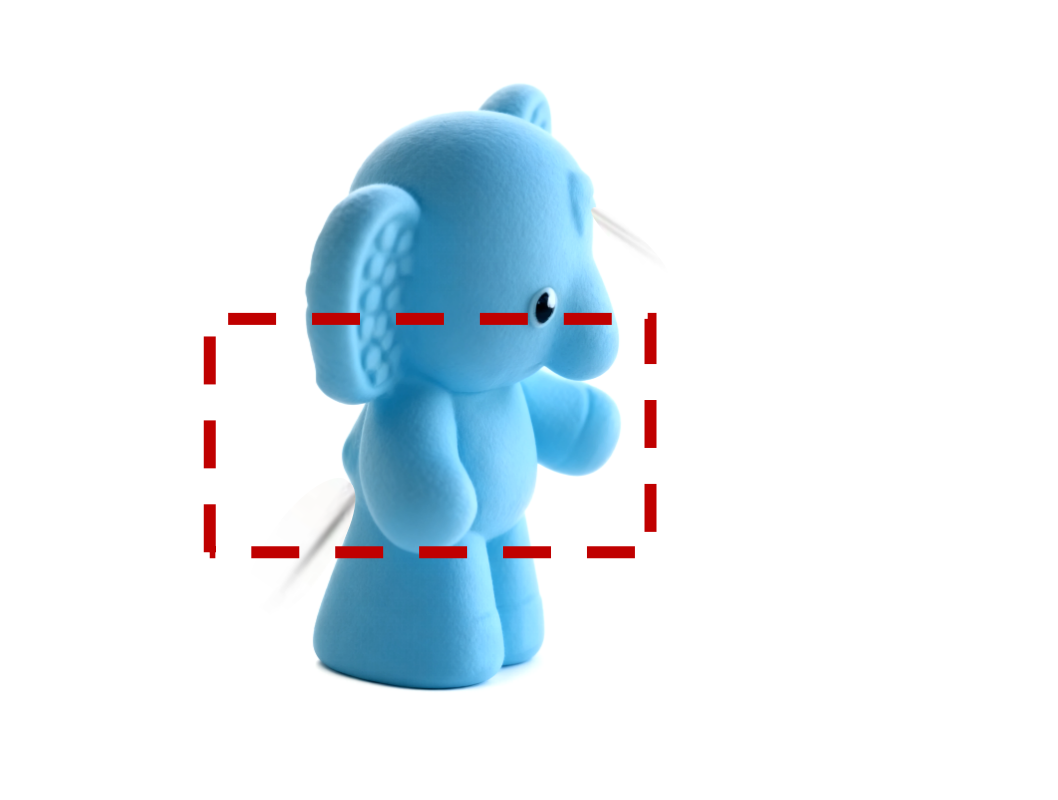} &
\includegraphics[width=.195\linewidth]{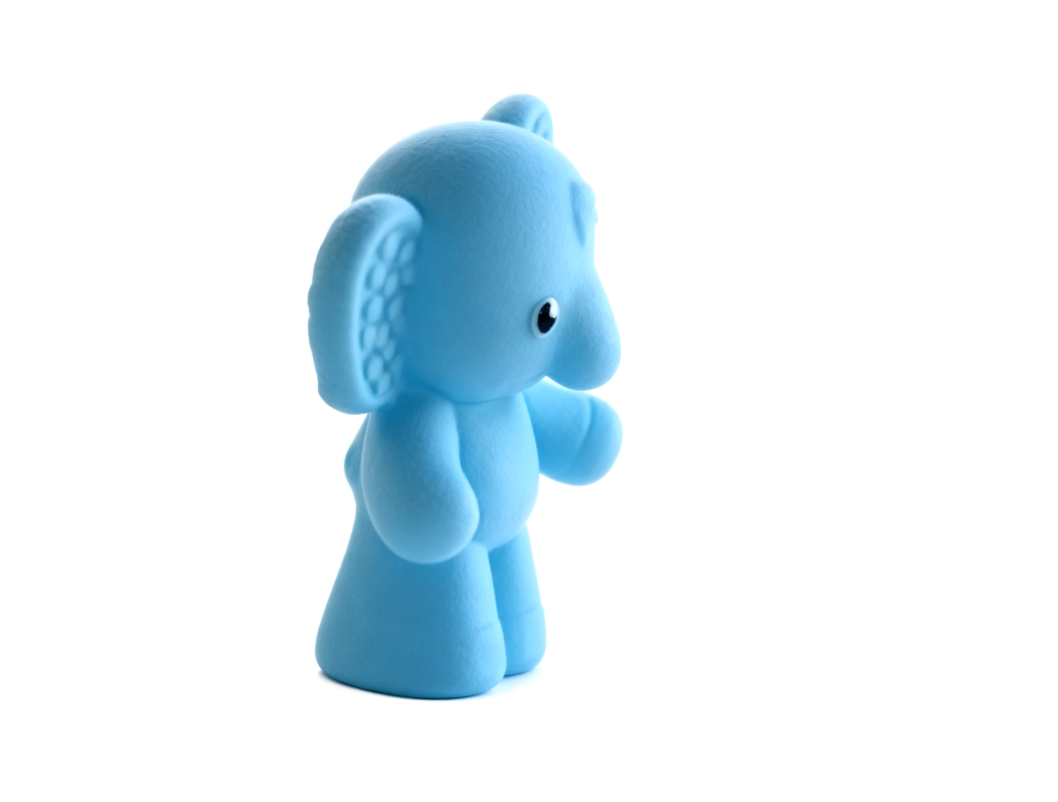} \\
\rotatebox{90}{\hspace{0.6cm} teatime} &
\includegraphics[width=.195\linewidth]{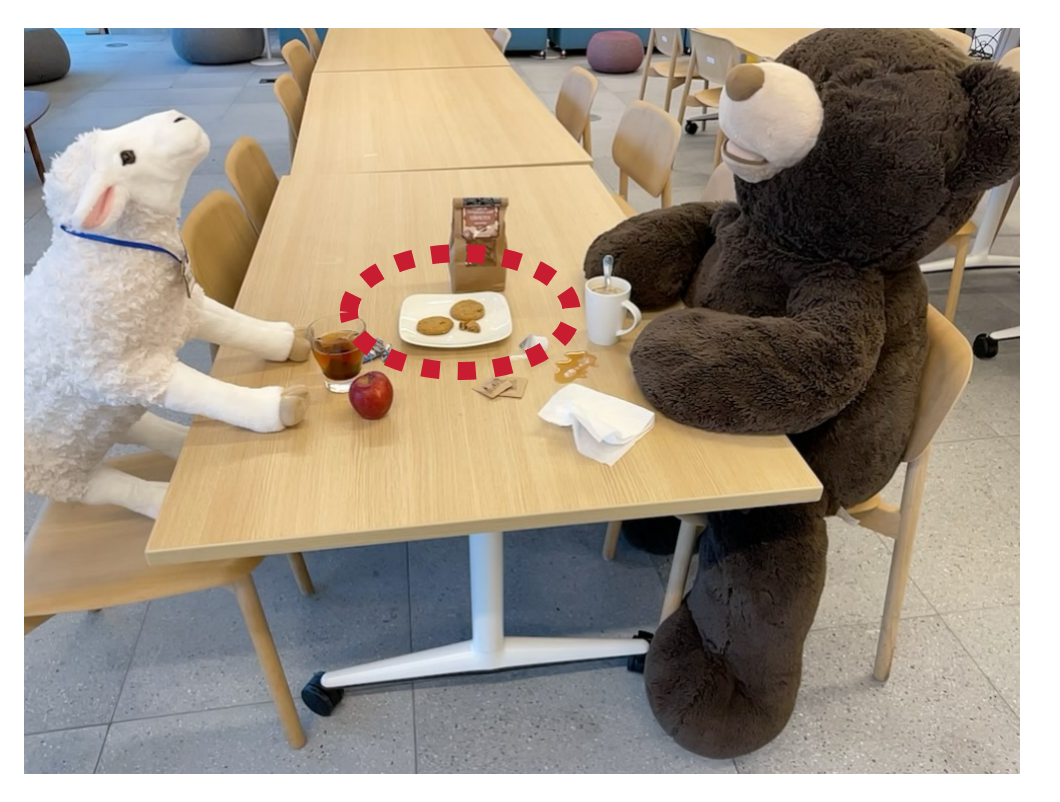} &
\includegraphics[width=.195\linewidth]{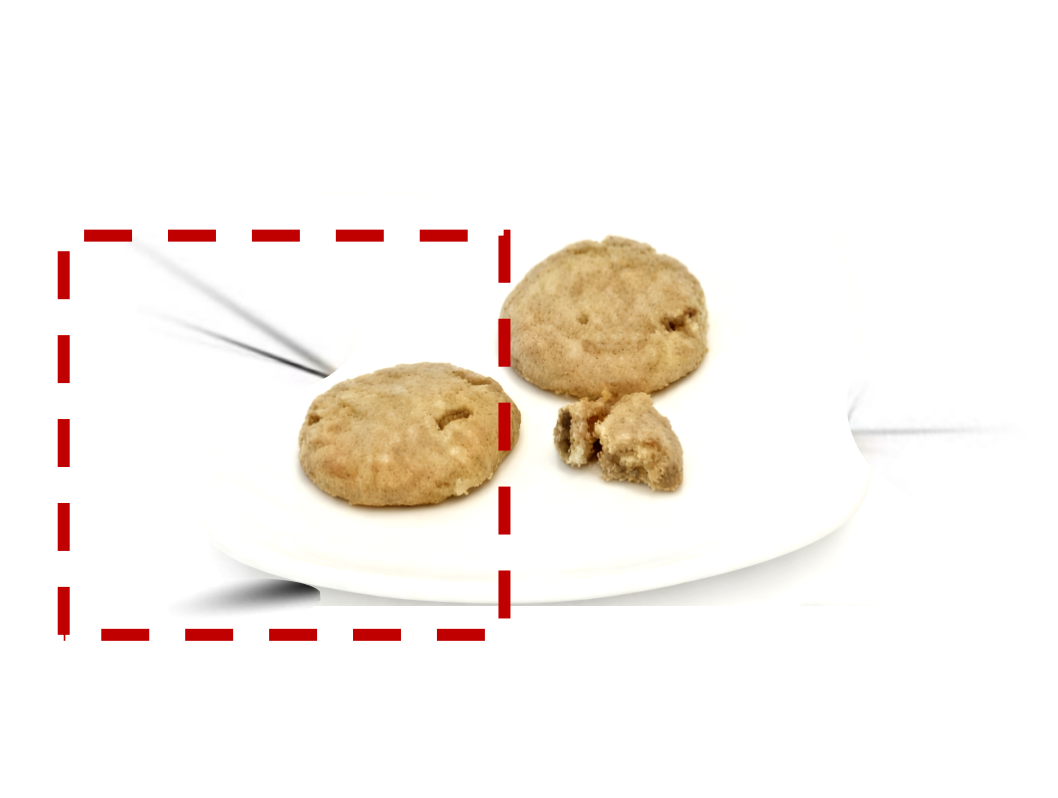} &
\includegraphics[width=.195\linewidth]{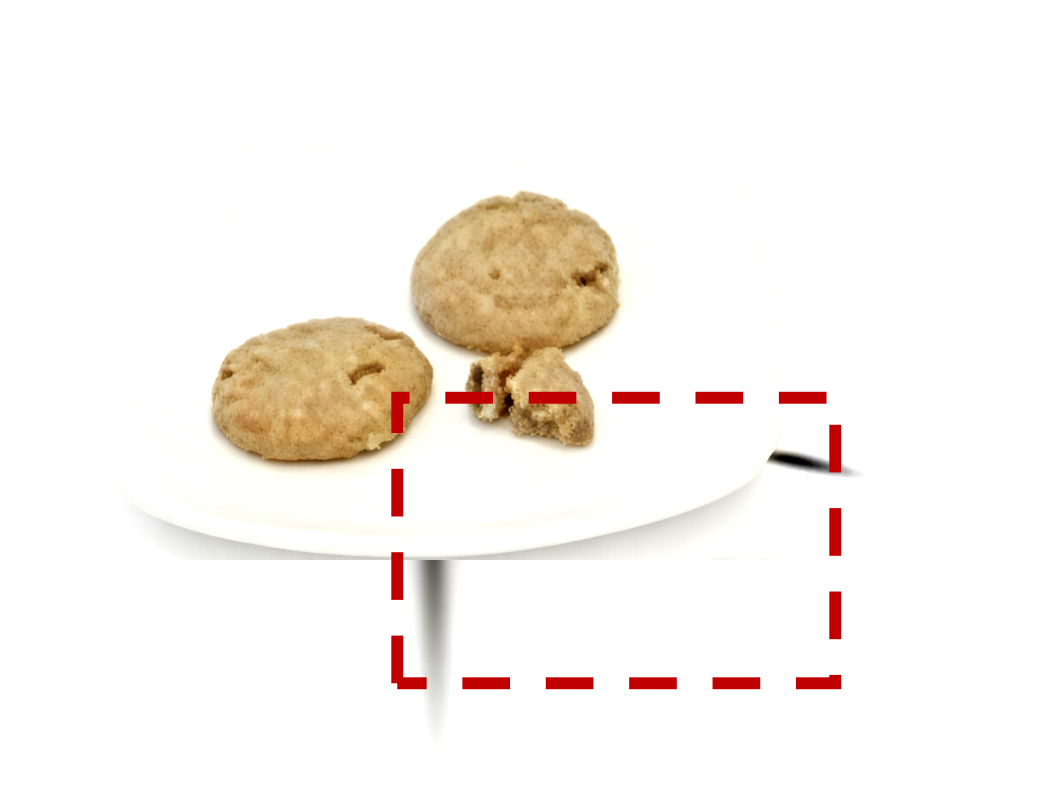} &
\includegraphics[width=.195\linewidth]{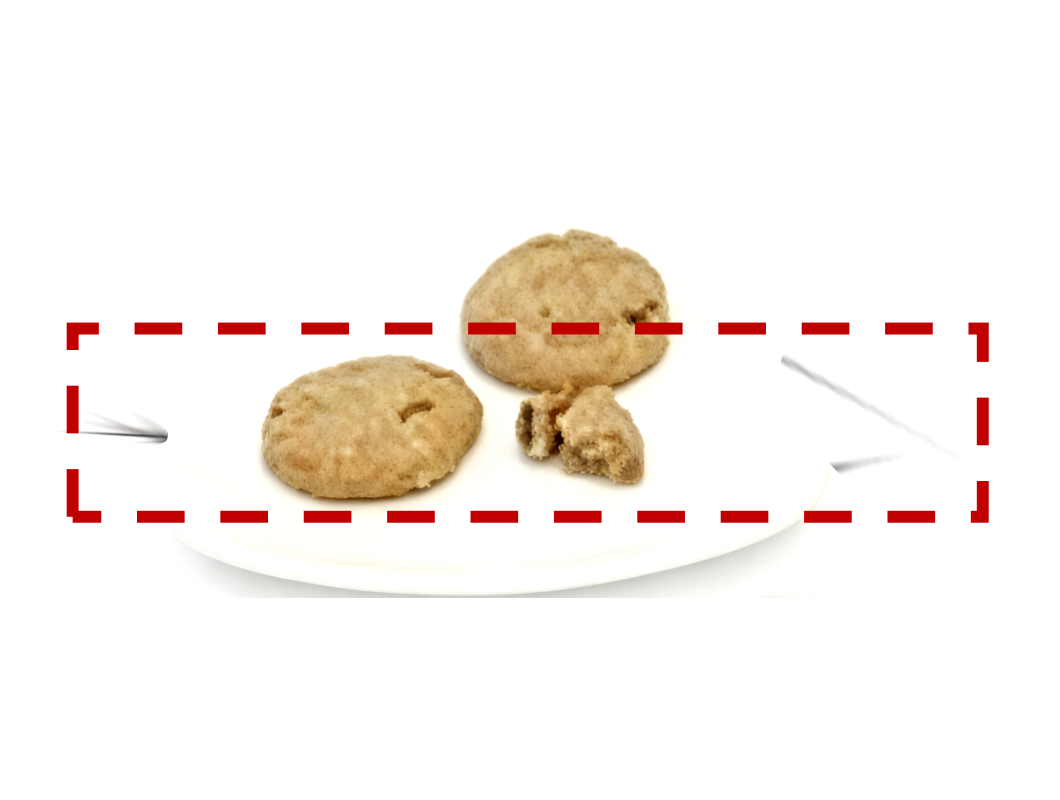} &
\includegraphics[width=.195\linewidth]{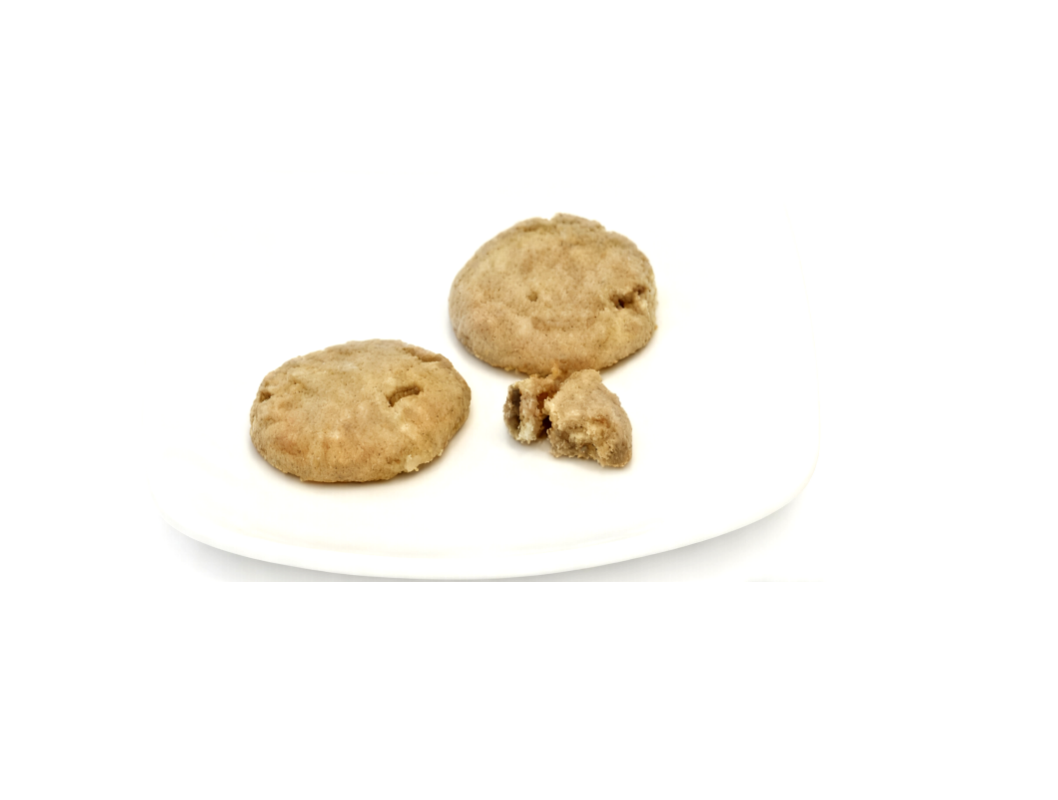} \\
& (a) Original scene & (b) SAGA & (c) FlashSplat & (d) COB-GS & (e) Our method \\
\end{tabular}
\caption{Qualitative result on NVOS and LERF-OVS datasets. The results show that our method segments the boundaries of the object more clearly, without blurred Gaussians.}
\label{fig:samples}
\end{figure*}

\subsection{Qualitative Results}

Figure~\ref{fig:samples} shows qualitative comparison of segmentation results on the NVOS and LERF-OVS datasets, showing the performance of our method compared to three best-performing baselines (SAGA, FlashSplat, and COB-GS). In the scene \textit{fern}, it accurately preserves the fine structures of leaves and stems, whereas SAGA and FlashSplat produce fragmented results, and COB-GS introduces background artifacts. For \textit{horn}, our method maintains the integrity of complex bone details, while SAGA oversmooths and FlashSplat yields incomplete boundaries. In \textit{figurines} and \textit{teatime}, our method successfully captures object contours (e.g., elephant, plate) and natural shape transitions, while others struggle with blurred boundaries, especially in fine structural areas. Red bounding boxes highlight key areas where our method has achieved significant improvements in boundary segmentation and spatial continuity.

\subsection{Computational Efficiency Analysis}
Table~\ref{tab:time consumption} shows the time consumption of our method in comparison with the baseline methods in the \textit{fortress} scene. 
Feedforward methods utilize pre-trained models and do not need optimization for each scene. However, their segmentation accuracy is limited for complex scenes.
Mask-based methods are trained for each scene to optimize the 3D Gaussian distribution. Our method achieves similar computational efficiency to COB-GS and outperforms other mask-based methods in both training and inference efficiency. 

\begin{table}[!t]
\setlength{\tabcolsep}{8.pt} 
\centering
\caption{Time consumption comparison in the \textit{fortress} scene.}
\label{tab:performance_comparison_overall}
\begin{tabular}{lcc}
\toprule
\textbf{Method} & \textbf{Training (m)} & \textbf{Inference (s)}  \\
\midrule
LSeg~\cite{liu2023lseg} & - &   1.6\\
LangSplat~\cite{tang2023langsplat} & - &   1.2\\
LangSurf~\cite{chen2024langsurf} & - &   1.1\\
LSM~\cite{yang2024lsm} & - &  0.9\\
LanScene-X~\cite{zhang2024lanscene} & - &  0.9\\
\hline
SA3D-GS~\cite{cen2025segment} & 15 & 1.5 \\
SAGA~\cite{wang2024saga} & 13 & 1.2  \\
FlashSplat~\cite{chen2024flashsplat} & 12 & 1.0  \\
COB-GS~\cite{li2024cobgs} &  10& 0.8  \\
\hline
Ours & \textbf{9} & \textbf{0.8} \\
\bottomrule
\end{tabular}
\vspace{-0.5cm}
\label{tab:time consumption}
\end{table}

\subsection{Ablation Studies}

\begin{table}[!h]
\vspace{-0.1cm}
\small
\setlength{\tabcolsep}{1.2pt} 
\centering
\caption{Ablation study of different components on NVOS dataset.}
\label{tab:ablation}
\begin{tabular}{lcccccccccc}
\toprule
\multicolumn{7}{c}{\textbf{Components}} & \multicolumn{3}{c}{\textbf{Performance}} \\
\cmidrule(lr){1-7} \cmidrule(lr){8-10}
Bd. & Sp. & RBF. & MRHE. & NeRF. & $\mathcal{L}_{\text{align}}$. & $\mathcal{L}_{\text{smth}}$. & B-mIoU & mAcc & mIoU \\
\midrule
$\checkmark$ & $\checkmark$ & $\checkmark$ & $\checkmark$ & $\checkmark$ & $\checkmark$ & $\checkmark$ & \textbf{84.7} & \textbf{99.2} & \textbf{92.6} \\
$\checkmark$ & $\checkmark$ & $\checkmark$ & $\checkmark$ & $\checkmark$ & $\checkmark$ & - & 84.1 & 99.1 & 92.5 \\
$\checkmark$ & $\checkmark$ & $\checkmark$ & $\checkmark$ & $\checkmark$ & - & - & 83.4 & 99.1 & 92.4 \\
$\checkmark$ & $\checkmark$ & $\checkmark$ & $\checkmark$ & - & - & - & 82.8 & 99.0 & 92.4 \\
$\checkmark$ & $\checkmark$ & $\checkmark$ & - & - & - & - & 82.2 & 99.0 & 92.4 \\
$\checkmark$ & $\checkmark$ & - & - & - & - & - & 80.5 & 98.5 & 92.0 \\
$\checkmark$ & - & - & - & - & - & - & 79.0 & 98.3 & 91.8 \\
- & - & - & - & - & - & - & 77.5 & 98.2 & 91.6 \\
\bottomrule
\end{tabular}
\vspace{-0.1cm}
\end{table}

To verify the contribution of each module in our method, we present the ablation study results in Table~\ref{tab:ablation}. It shows the performance changes on the NOVS dataset when different components are gradually removed from the original network. Removing ($\mathcal{L}_{\text{align}}$ and $\mathcal{L}_{\text{smth}}$) reduces B-mIoU by 1.3\%. Further removing the NeRF module, the MRHE encoding, and finally the RBF interpolation causes additional decreases of 0.7\%, 0.6\%, and 1.7\% in B-mIoU, respectively. Notably, removing the boundary-region sampling (Sp) alone causes a substantial decline of 1.5\%. Finally, removing the boundary-Gaussian selection (Bd) further lowers the performance to 77.5\%. The results underscore the critical role of RBF interpolation in providing continuous feature representation, which effectively regulates the NeRF continuous modeling through interpolation features and enhances boundary consistency.

\begin{figure}[t]
    \centering
    \begin{subfigure}[b]{0.1558\textwidth}
        \includegraphics[width=\textwidth]{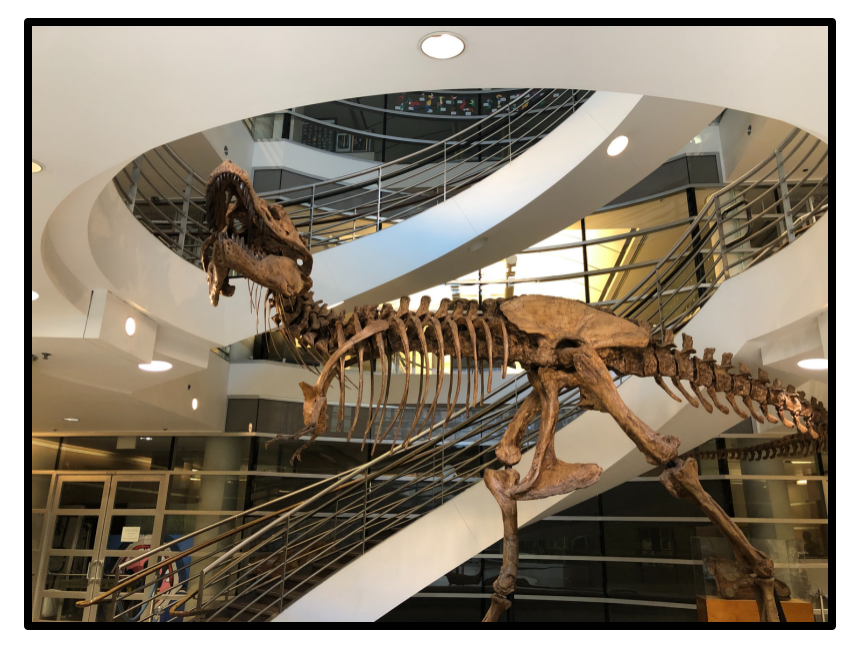}
       
        \label{fig:boundary_smoothness}
    \end{subfigure}
    \hfill
    \begin{subfigure}[b]{0.1558\textwidth}
        \includegraphics[width=\textwidth]{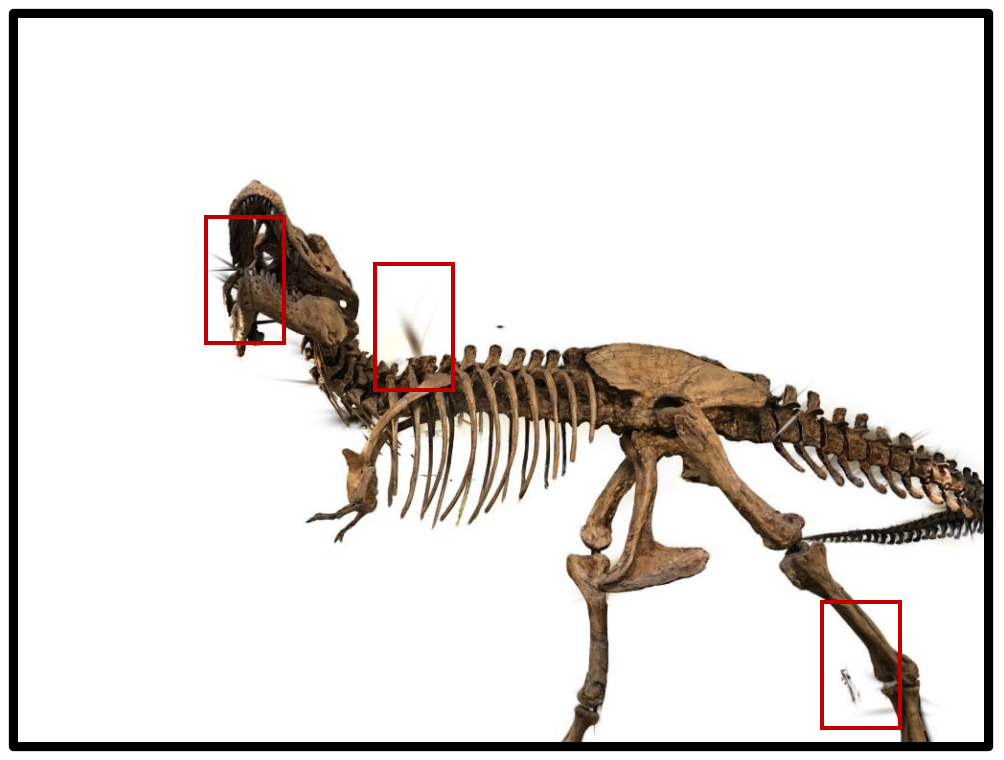}
        
        \label{fig:novs_forns_center}
    \end{subfigure}
    \hfill
    \begin{subfigure}[b]{0.1558\textwidth}
        \includegraphics[width=\textwidth]{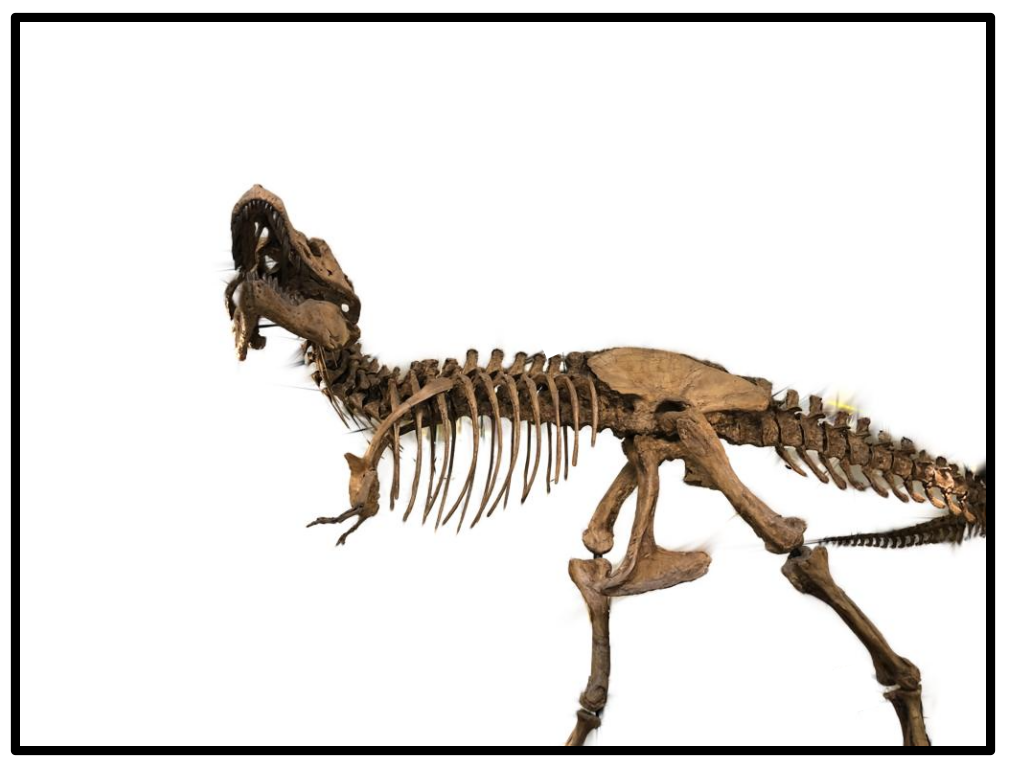}
        
        \label{fig:mutated_gaussian}
    \end{subfigure}
    \hfill
    \begin{subfigure}[b]{0.1558\textwidth}
        \includegraphics[width=\textwidth]{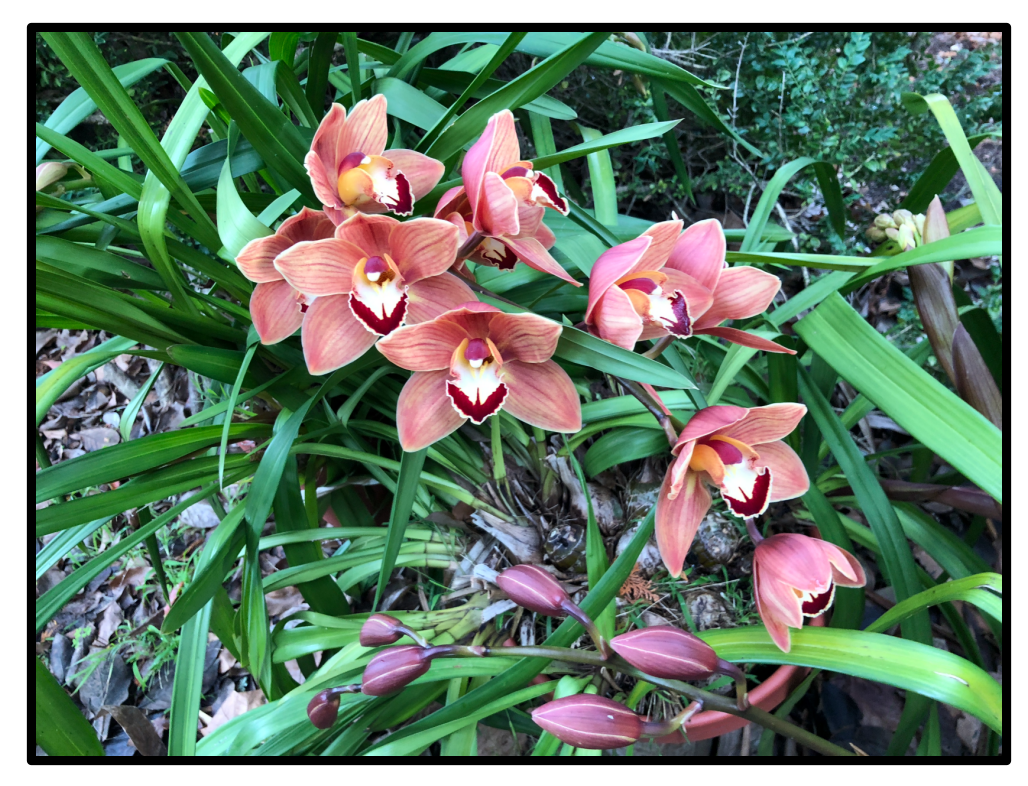}
        \caption{Original scene}
        \label{fig:novs_forns_center}
    \end{subfigure}
    \hfill
    \begin{subfigure}[b]{0.1558\textwidth}
        \includegraphics[width=\textwidth]{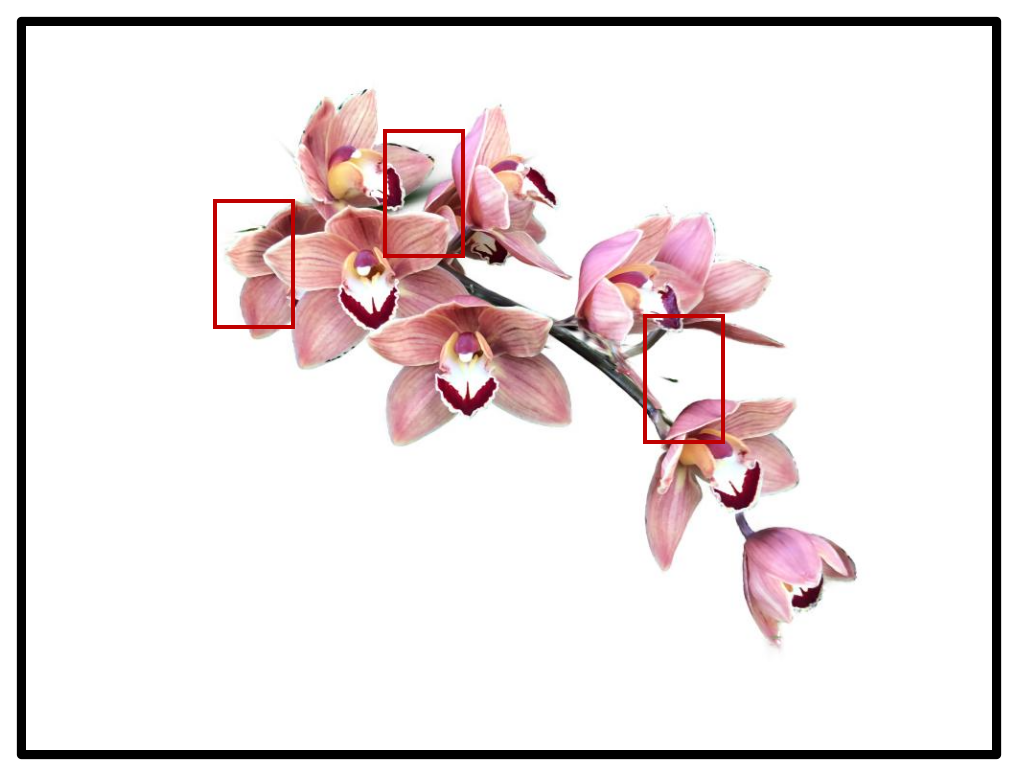}
        \caption{w/o RBF+MRHE}
        \label{fig:mutated_gaussian}
    \end{subfigure}
    \hfill
    \begin{subfigure}[b]{0.1558\textwidth}
        \includegraphics[width=\textwidth]{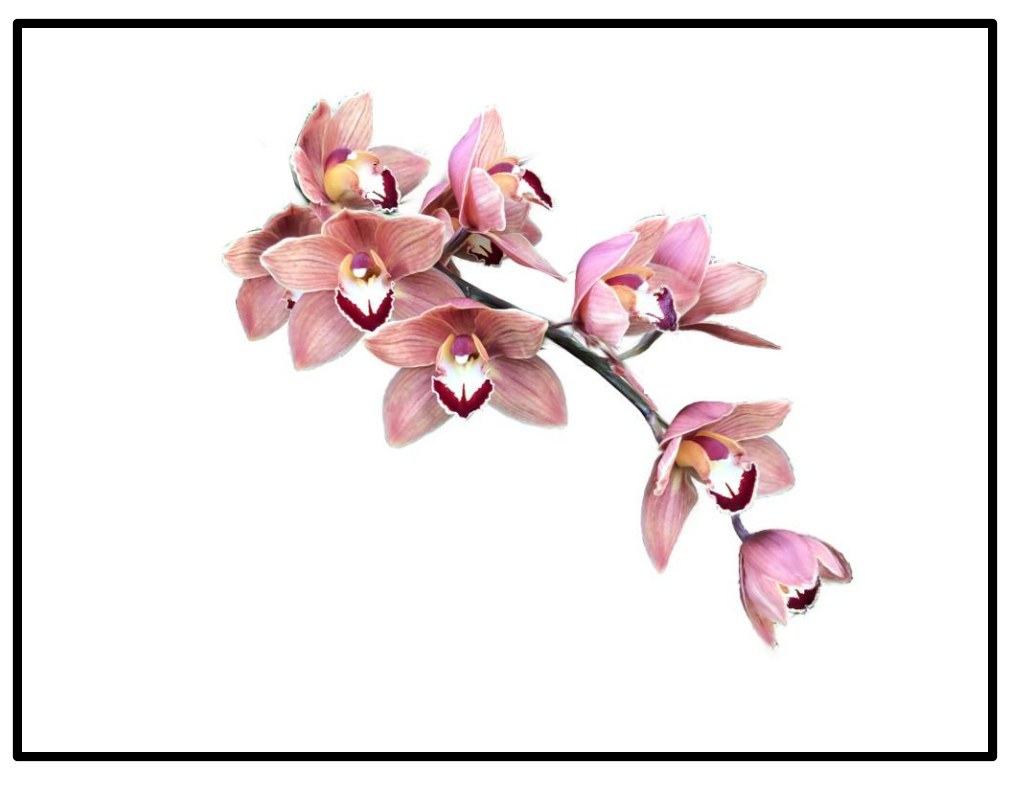}
        \caption{w/ RBF+MRHE}
        \label{fig:our_method}
    \end{subfigure}
    
    \caption{Ablation results in \textit{trex} and \textit{orchid} scenes.}
\label{fig:ablation}
\vspace{-0.4cm}
\end{figure}

A qualitative comparison in Figure~\ref{fig:ablation} between our method and reduced (without RBF+MRHE) methods on the \textit{trex} and \textit{orchid} scenes visually confirms this analysis. The reduced method produces fragmented segmentation with boundary artifacts, disrupting structural continuity in fine skeletal details (\textit{trex}) and failing to capture subtle petal boundaries (\textit{orchid}). In contrast, our method generates coherent boundaries and maintains accurate shapes across both complex structures.

\begin{figure}[t]
    \centering
    \begin{subfigure}[b]{0.115\textwidth}
        \includegraphics[width=\textwidth]{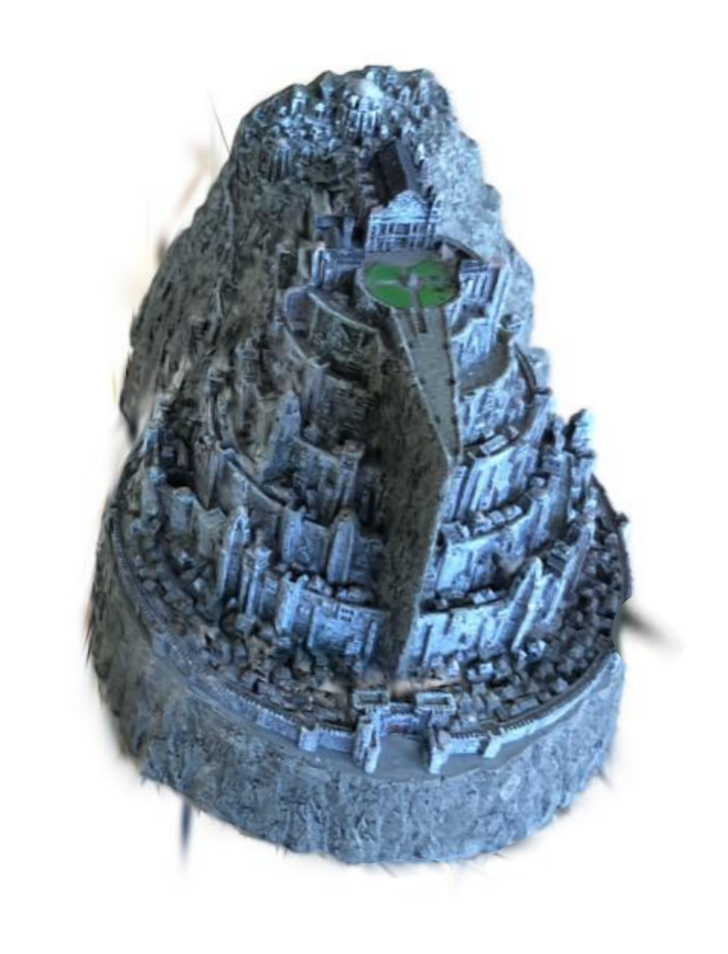}
        \caption{$\tau$=0.9}
        \label{fig:boundary_smoothness}
    \end{subfigure}
    \hfill
    \begin{subfigure}[b]{0.11\textwidth}
        \includegraphics[width=\textwidth]{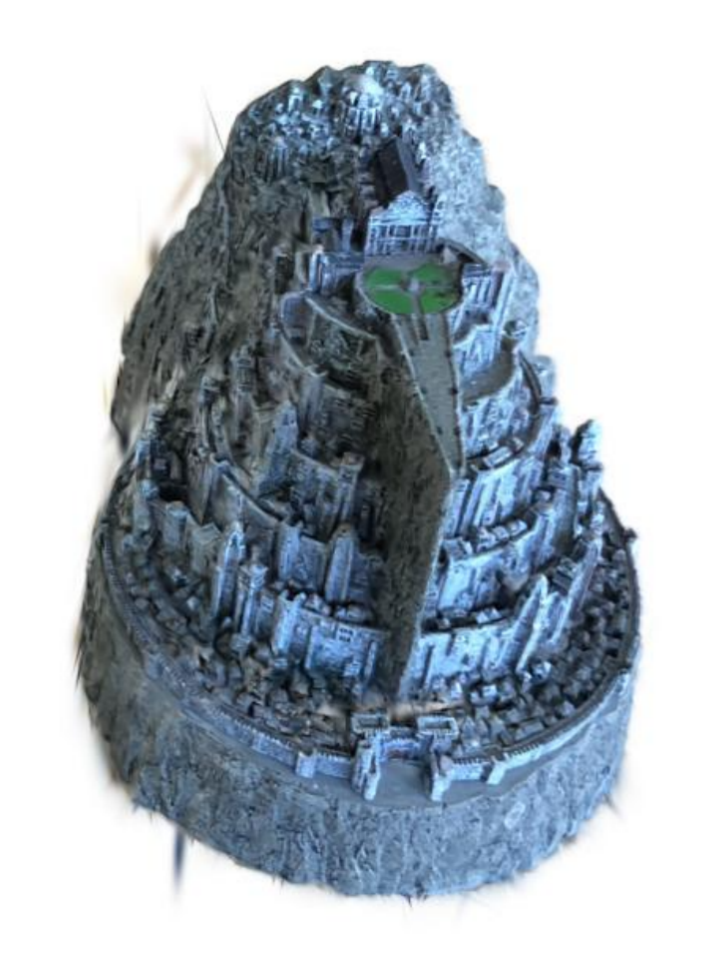}
        \caption{$\tau$=0.3}
        \label{fig:novs_forns_center}
    \end{subfigure}
    \hfill
    \begin{subfigure}[b]{0.11\textwidth}
        \includegraphics[width=\textwidth]{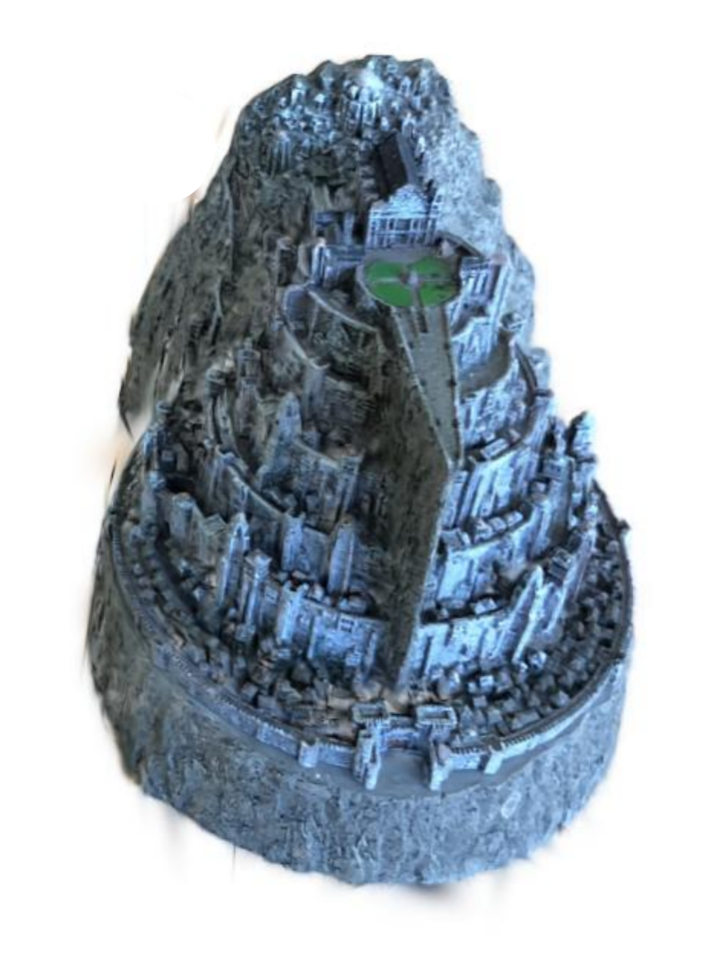}
        \caption{$\tau$=0.7}
        \label{fig:mutated_gaussian}
    \end{subfigure}
    \hfill
    \begin{subfigure}[b]{0.11\textwidth}
        \includegraphics[width=\textwidth]{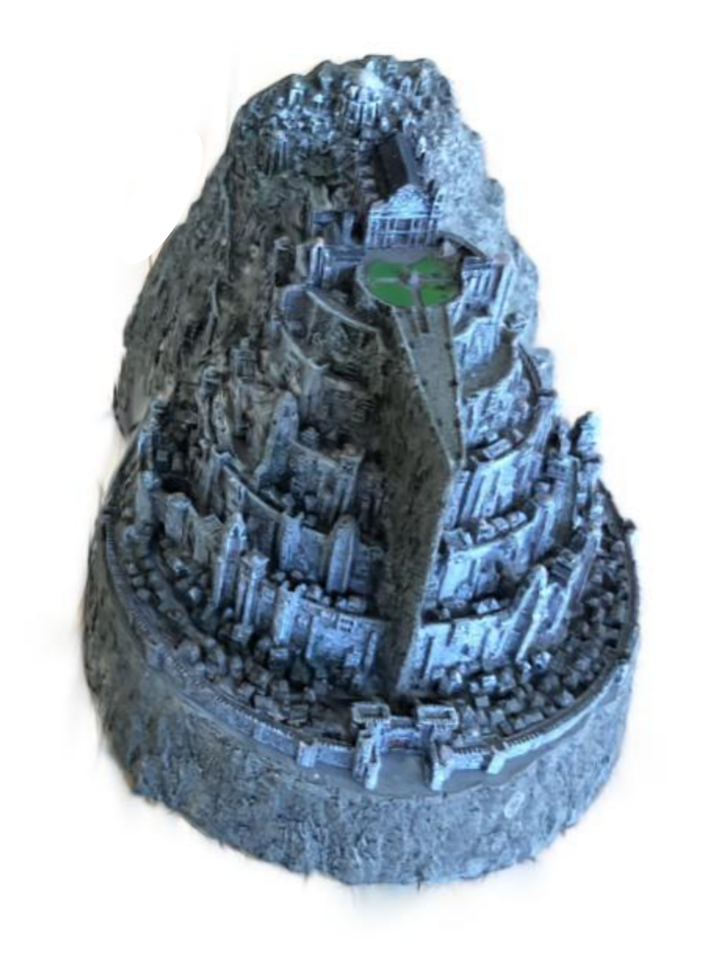}
        \caption{$\tau$=0.6}
        \label{fig:our_method}
    \end{subfigure}
    
    \caption{The influence of segmentation threshold ($\tau$).}
    \label{fig:tau}
\vspace{-0.5cm}
\end{figure}

\subsection{Hyper-parameter Analysis}
Here we investigate the impact of the segmentation threshold $\tau$ (Eq.~\ref{eq:boundary}) in both quantitative performance and visual quality. Figure~\ref{fig:tau} displays the segmentation results under four thresholds ($\tau$=0.9, 0.3, 0.7, 0.6) in the fortress scene. It is shown that $\tau$=0.6 achieves the best balance between maintaining structural integrity and controlling background noise, resulting in excellent visual coherence and detail preservation. The results presented in Table~\ref{tab:tau} quantitatively confirm the visual finding, where $\tau$=0.6 produces peak performance in both mIoU and B-mIoU. 

\begin{table}[!t]
\setlength{\tabcolsep}{10pt} 
\centering

\caption{Segmentation results with different thresholds $\tau$.}
\label{tab:tau_iou}
\begin{tabular}{cccccc}
\toprule
$\tau$ & 0.3  & 0.6 & 0.7 & 0.9 \\
\midrule
mIoU (\%) & 65.2 & 92.6 & 91.1 & 54.6 \\
B-mIoU(\%) &57.3& 84.7& 81.4& 51.2 \\
\bottomrule
\end{tabular}
\label{tab:tau}
\vspace{-0.5cm}
\end{table}

\section{Conclusion}
In this paper, we presented NG-GS, a novel framework for accurate and continuous 3DGS segmentation. By leveraging variance-based boundary detection, RBF-based feature interpolation, and a joint NeRF-3DGS optimization strategy, NG-GS effectively mitigates the discretization artifacts commonly encountered around object boundaries. Extensive experiments validate that NG-GS outperforms compared baseline methods in both quantitative and qualitative evaluations, particularly in boundary-aware metrics. Our work underscores the importance of continuity modeling in explicit 3D representations and opens up new possibilities for high-precision scene understanding and manipulation. 

\vspace{.5mm}\noindent\textbf{Future work/Limitation.} Addressing current limitations, our future directions include extending the framework to dynamic scenes and real-time interactive applications, further bridging the gap between representation learning and practical vision systems.   
\section*{Acknowledgments}
This work is supported by the Fundamental Research Funds for the Central Universities (No. 2025JBZY011), the National Nature Science Foundation of China (Nos. 62376020 and 62571027) and the Beijing Natural Science-Xiaomi Innovation Joint Foundation (No. L253007).

\input{}
{
    \small
    \bibliographystyle{ieeenat_fullname}
    \bibliography{main}
}
\clearpage
\setcounter{page}{1}
\maketitlesupplementary

The supplementary material provides complete implementation details and extended analyses. Specifically, it includes: the Edge Gaussian Continuity algorithm (Sec.~\ref{Edge Gaussian Continuity}), a computation cost analysis comparing efficiency (Sec.~\ref{Computation Cost}), an open-vocabulary segmentation evaluation (Sec.~\ref{Open-Vocabulary}, Table~\ref{tab:open}), a test of robustness to erroneous masks (Sec.~\ref{Robustness against erroneous masks}), and a hyperparameter analysis (Sec.~\ref{Hyperparameter experiment}). We provide expanded quantitative results: a computation cost comparison (Table~\ref{tab:time_comparison}), per-scene results on LERF-OVS (Table~\ref{tab:open}), and CLIP-IQA scores (Table~\ref{tab:clip}).

\section{Edge Gaussian Continuity}
\label{Edge Gaussian Continuity}
The Edge Gaussian Continuity algorithm is designed to address the discrete boundary issues in 3DGS by constructing a spatially continuous feature field. It begins by identifying boundary Gaussians through variance analysis of multi-view 2D masks, where Gaussians with high variance values are selected as ambiguous boundary points. These points are then projected onto a reference image plane to compute an expanded bounding box. Dense grid sampling is performed within this box to generate 2D sample points, which are subsequently used to create 3D query points along camera rays via depth-adaptive sampling. For each query point, RBF interpolation is applied using K-nearest neighbors from the boundary Gaussians, with weights based on Gaussian kernel distances. Finally, MRHE enhances the interpolated features by capturing multi-scale spatial information, which improves boundary smoothness and segmentation consistency in the NG-GS framework. See Algorithm 1 for details.

\section{Computation Cost}
\label{Computation Cost}

We compare the computational efficiency of NG-GS with state-of-the-art 3DGS based methods and feedforward based methods. This evaluation was conducted using a single NVIDIA RTX 3090 GPU on all scenarios from the NOVS dataset ~\cite{ren2022neural}, and the results are shown in Table ~\ref{tab:time_comparison}. We provide the average total training time and inference time for the entire reconstruction and segmentation pipeline. Compared with COB-GS, our segmentation process does not rely on edge Gaussian splitting to remove mutated Gaussian, but utilizes NeRF and fast modeling MRHE, which ensures edge optimization while optimizing scene labels. The optimization time is comparable to the speed of COB-GS.

\begin{algorithm}[H]
    \caption{Edge Gaussian Continuity}
    \label{alg:edge_gaussian_continuity}
     \begin{algorithmic}
        \STATE \textbf{Input}: Trained 3DGS model $\mathcal{G}$, 2D mask set $M$, camera parameters, variance threshold $\tau$, grid size $N_{row}$, $N_{col}$, sampling depth $K$, kernel width $\sigma$, expansion offset $\delta$
        \STATE \textbf{Result}: Spatially continuous feature field
        
        \STATE // Step 1: Identify boundary Gaussians
        \FOR{each Gaussian $g_i$ in $\mathcal{G}$}
            \STATE $v_i \gets \text{ComputeMaskVariance}(M, g_i)$
            \IF{$v_i > \tau$}
                \STATE $\mathcal{B}[g_i] \gets g_i$
            \ENDIF
        \ENDFOR
        
        \STATE // Step 2: Project and compute bounding box
        \STATE Project all $\mathcal{B}$ to image coordinates for a reference view
        \STATE $B_{min} \gets \min(\text{coordinates}) - \delta$
        \STATE $B_{max} \gets \max(\text{coordinates}) + \delta$
        
        \STATE // Step 3: Dense grid sampling
        \STATE $\Delta x \gets (B_{max}.x - B_{min}.x) / N_{col}$
        \STATE $\Delta y \gets (B_{max}.y - B_{min}.y) / N_{row}$
        \FOR{$k \gets 0$ to $N_{col}$}
            \FOR{$l \gets 0$ to $N_{row}$}
                \STATE $p \gets (B_{min}.x + k \cdot \Delta x, B_{min}.y + l \cdot \Delta y)$
                \STATE $\mathcal{P}_{grid} \gets \mathcal{P}_{grid} \cup \{p\}$
            \ENDFOR
        \ENDFOR
        
        \STATE // Step 4: Generate query points along rays
        \FOR{each sample point $p_i$ in $\mathcal{P}_{grid}$}
            \STATE $d_{world} \gets \text{ComputeRayDirection}(p_i, \text{camera params})$
            \FOR{$s \gets 1$ to $K$}
                \STATE $\delta_s \gets -1 + 2 \cdot (s-1)/(K-1)$
                \STATE $q \gets o + [\bar{t} + \max(\alpha \cdot \bar{t}, \varepsilon) \cdot \delta_s] \cdot d_{world}$
                \STATE $\mathcal{P}_{query} \gets \mathcal{P}_{query} \cup \{q\}$
            \ENDFOR
        \ENDFOR
        
        \STATE // Step 5: RBF interpolation
        \FOR{each query point $q$ in $\mathcal{P}_{query}$}
            \STATE $\mathcal{N}_q \gets \text{KNN}(\mathcal{B}, q)$
            \FOR{each Gaussian $g_j$ in $\mathcal{N}_q$}
                \STATE $w_j \gets \exp(-\|q - x_j\|^2 / (2\sigma^2))$
            \ENDFOR
            \STATE Normalize weights $w_j$ to sum to 1
            \STATE $f^{inter}(q) \gets \sum_{g_j \in \mathcal{N}_q} w_j \cdot f_j$
        \ENDFOR
        
        \STATE // Step 6: Multi-resolution hash encoding
        \FOR{each query point $q$ in $\mathcal{P}_{query}$}
            \STATE $f^{hash} \gets \text{MRHE}(q)$
        \ENDFOR
    \end{algorithmic}
\end{algorithm}

\label{sec:rationale}

\begin{figure*}[!t]
\centering
\includegraphics[height=0.31\textheight,width=0.89\textwidth,trim=0.4cm 0.5cm 0.5cm 0.3cm,clip]{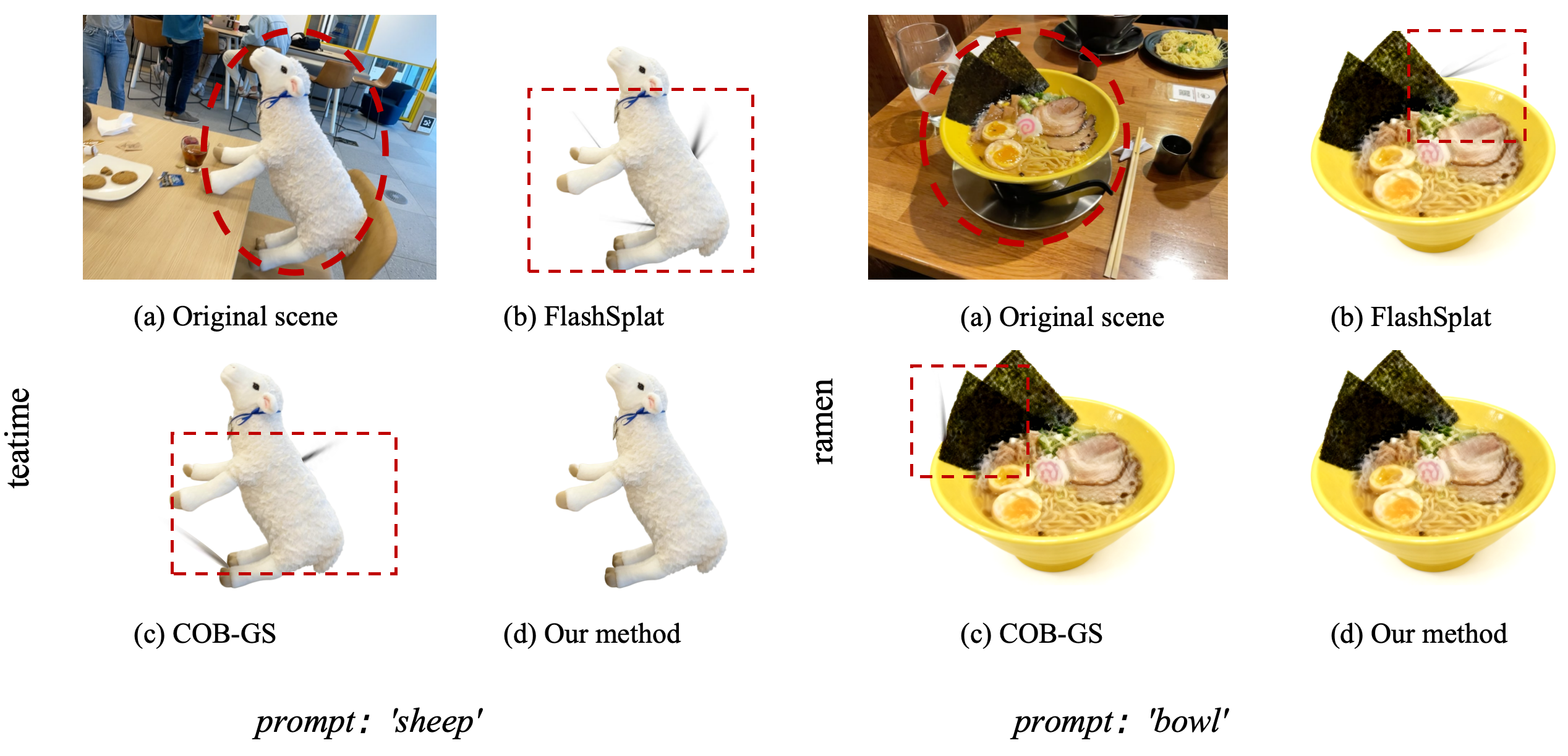}
  \caption{More qualitative comparisons of open-vocabulary 3D Gaussian Splatting segmentation on LERF-OVS dataset.}
  \label{fig:8}
\end{figure*}

\begin{table*}[!t]
\centering
\caption{Average training and inference times consumption comparison across all scenes.}
\label{tab:time_comparison}
\setlength{\tabcolsep}{1.1pt} 
\begin{tabular}{lccccc|ccccc}
\toprule
\textbf{Metric} & LSeg & LangSplat & LangSurf & LSM & LanScene-X~ & ~SA3D-GS & SAGA & FlashSplat & COB-GS & Ours \\
\midrule
Training time (m) & -- & -- & -- & -- & --~ & ~17 & 14 & 15 & 13 & \textbf{12} \\
Inference time (s) & 1.8 & 1.5 & 1.3 & 1.4 & 1.2~ & ~1.4 & 1.2 & 1.1 & 1.0 & \textbf{0.9} \\
\bottomrule
\end{tabular}
\end{table*}

\begin{table*}[!t]
\setlength{\tabcolsep}{4.5pt} 
\centering
\caption{Quantitative results on LERF-OVS dataset comparing various methods across three different scenes.}
\label{tab:open}
\begin{tabular}{lcccccc}
\toprule
\multirow{2}{*}{\textbf{Method}} & \multicolumn{2}{c}{\textbf{Figurines}} & \multicolumn{2}{c}{\textbf{Ramen}} & \multicolumn{2}{c}{\textbf{Teatime}} \\
\cmidrule(lr){2-3} \cmidrule(lr){4-5} \cmidrule(lr){6-7}
 & B-mIoU (\%) & mIoU (\%) & B-mIoU (\%) & mIoU (\%) & B-mIoU (\%) & mIoU (\%) \\
\midrule
LSeg~\cite{liu2023lseg} & 45.2 & 48.3 & 42.8 & 51.1 & 25.3 & 30.5 \\
LangSplat~\cite{tang2023langsplat} & 11.9 & 13.8 & 10.2 & 15.0 & 12.3 & 15.8 \\
LangSurf~\cite{chen2024langsurf} & 15.2 & 18.9 & 13.4  & 21.7 & 14.7  & 18.8\\
LSM~\cite{yang2024lsm} & 18.1 & 21.1 & 14.8 & 23.0 & 13.8  & 19.6 \\
LanScene-X~\cite{zhang2024lanscene} &37.4 & 40.1 & 34.7 & 42.9 & 40.1 & 45.0 \\
\hline
SA3D-GS~\cite{cen2025segment} & 23.9 & 24.9 & 7.0 & 7.4  & 39.2& 42.5\\
SAGA~\cite{wang2024saga} & 45.2 & 48.5 & 37.2 & 41.2 & 55.3 & 60.1 \\
FlashSplat~\cite{chen2024flashsplat} & 50.5 & 52.8 & 44.7 & 50.4 & 65.6 & 69.5 \\
COB-GS~\cite{li2024cobgs} & 73.9 & 76.3 & 69.2 & 78.1 & 72.8 & 77.2 \\
\hline
Ours & \textbf{74.2} & \textbf{75.4} & \textbf{77.5} & \textbf{79.3} & \textbf{76.3} & \textbf{79.2} \\
\bottomrule
\end{tabular}
\begin{minipage}{\textwidth}
\footnotesize
\end{minipage}
\end{table*}

\begin{table*}[!t]
\centering
\vspace{-0.1cm}
\caption{Quantitative comparison of CLIP-IQA for mask-based 3DGS segmentation methods.}
\label{tab:clip}
\setlength{\tabcolsep}{2.5pt} 
\begin{tabular}{lccc}
\toprule
\multirow{2}{*}{\textbf{Method}} & \multicolumn{3}{c}{\textbf{CLIP-IQA}} \\
\cmidrule{2-4}
 & Clear / Unclear Boundary & Smooth / Noisy Boundary & Complete / Mutilated Object \\
\midrule
SA3D-GS ~\cite{cen2025segment}& 65.8 & 71.8 & 83.5 \\
SAGA ~\cite{wang2024saga}& 66.2 & 62.4 & 84.6 \\
FlashSplat ~\cite{chen2024flashsplat}& 62.6 & 64.4 & 82.9 \\
COB-GS ~\cite{li2024cobgs}& 68.2 & 73.1 & 85.9 \\
\hline
Ours & \textbf{72.6} & \textbf{75.4} & \textbf{88.3} \\  
\bottomrule  
\end{tabular}
\end{table*}

\section{Open-Vocabulary 3D Segmentation}
\label{Open-Vocabulary}
We use the COB-GS to implement open vocabulary semantic segmentation. Some objects in this dataset have severe occlusion, and we show more examples of open vocabulary 3D semantic segmentation on the LERF-OVS~\cite{kerr2023lerf} dataset in Figure ~\ref{fig:8}. We observed that the results generated by COB-GS cannot provide the exact shape of the query object and contain a lot of noise, while our method can accurately describe the shape of the object. These results demonstrate the effectiveness of our proposed NG-GS in object edge segmentation as shown in Table ~\ref{tab:open}. The CLIP-IQA test compared with COB-GS is provided in the Table ~\ref{tab:clip}.

\section{Robustness Against Erroneous Masks}
\label{Robustness against erroneous masks}
\begin{figure}[H]
\vspace{-0.25cm}
    \centering
    \begin{subfigure}[b]{0.115\textwidth}
        \includegraphics[width=\textwidth]{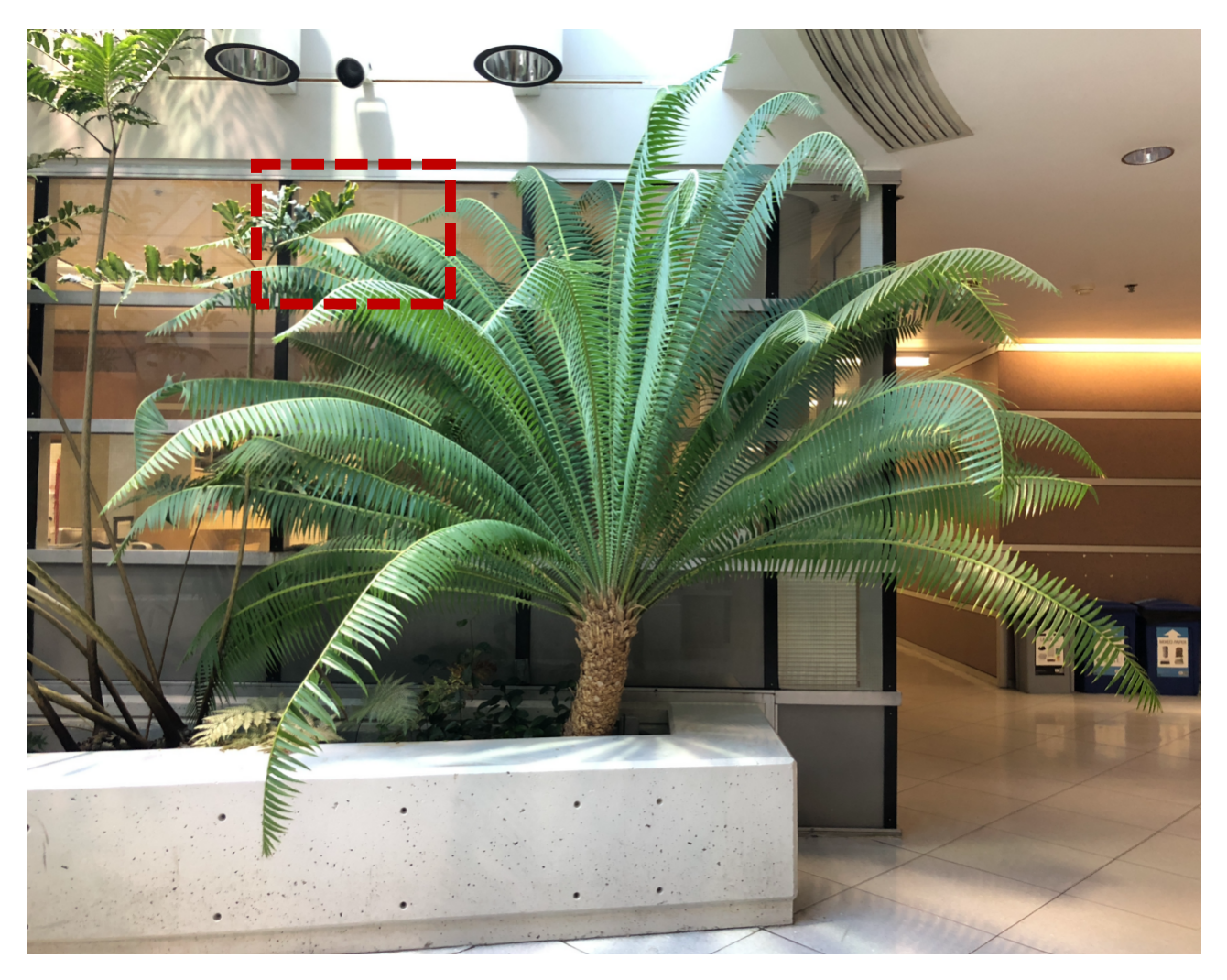}
        \caption{Input}
        \label{fig:boundary_smoothness}
    \end{subfigure}
    \hfill
    \begin{subfigure}[b]{0.115\textwidth}
        \includegraphics[width=\textwidth]{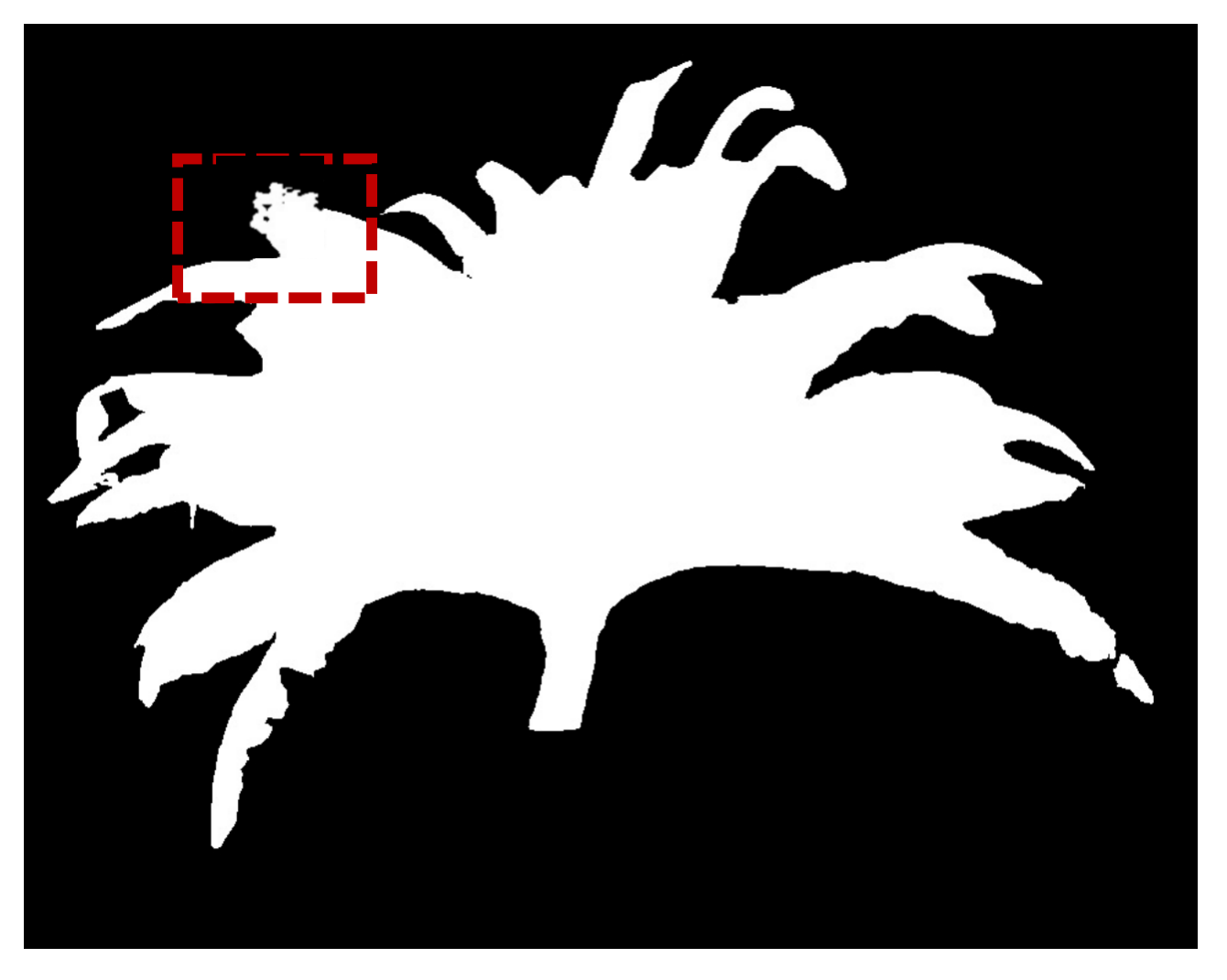}
        \caption{GT}
        \label{fig:novs_forns_center}
    \end{subfigure}
    \hfill
    \begin{subfigure}[b]{0.115\textwidth}
        \includegraphics[width=\textwidth]{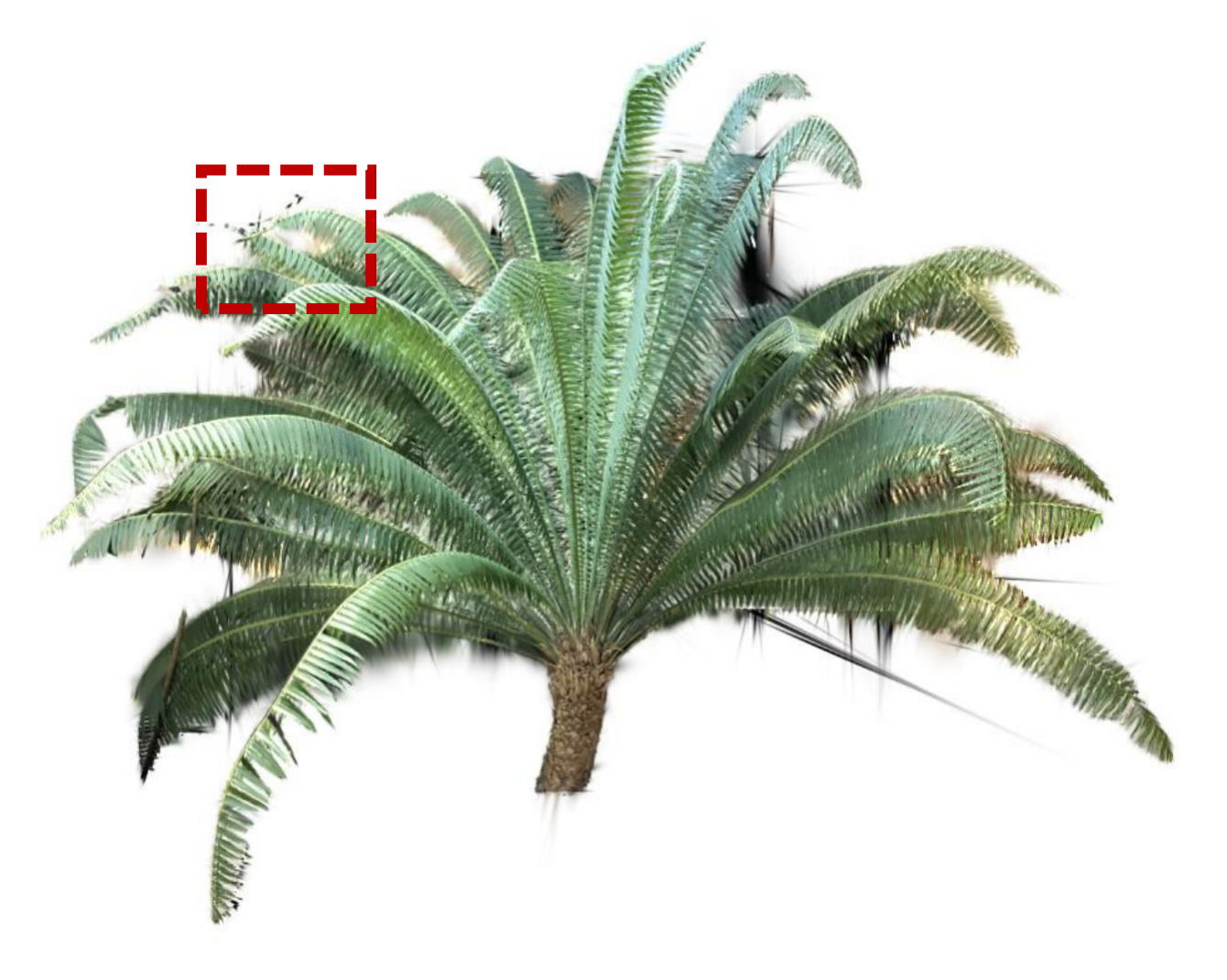}
        \caption{COB-GS}
        \label{fig:mutated_gaussian}
    \end{subfigure}
    \hfill
    \begin{subfigure}[b]{0.115\textwidth}
        \includegraphics[width=\textwidth]{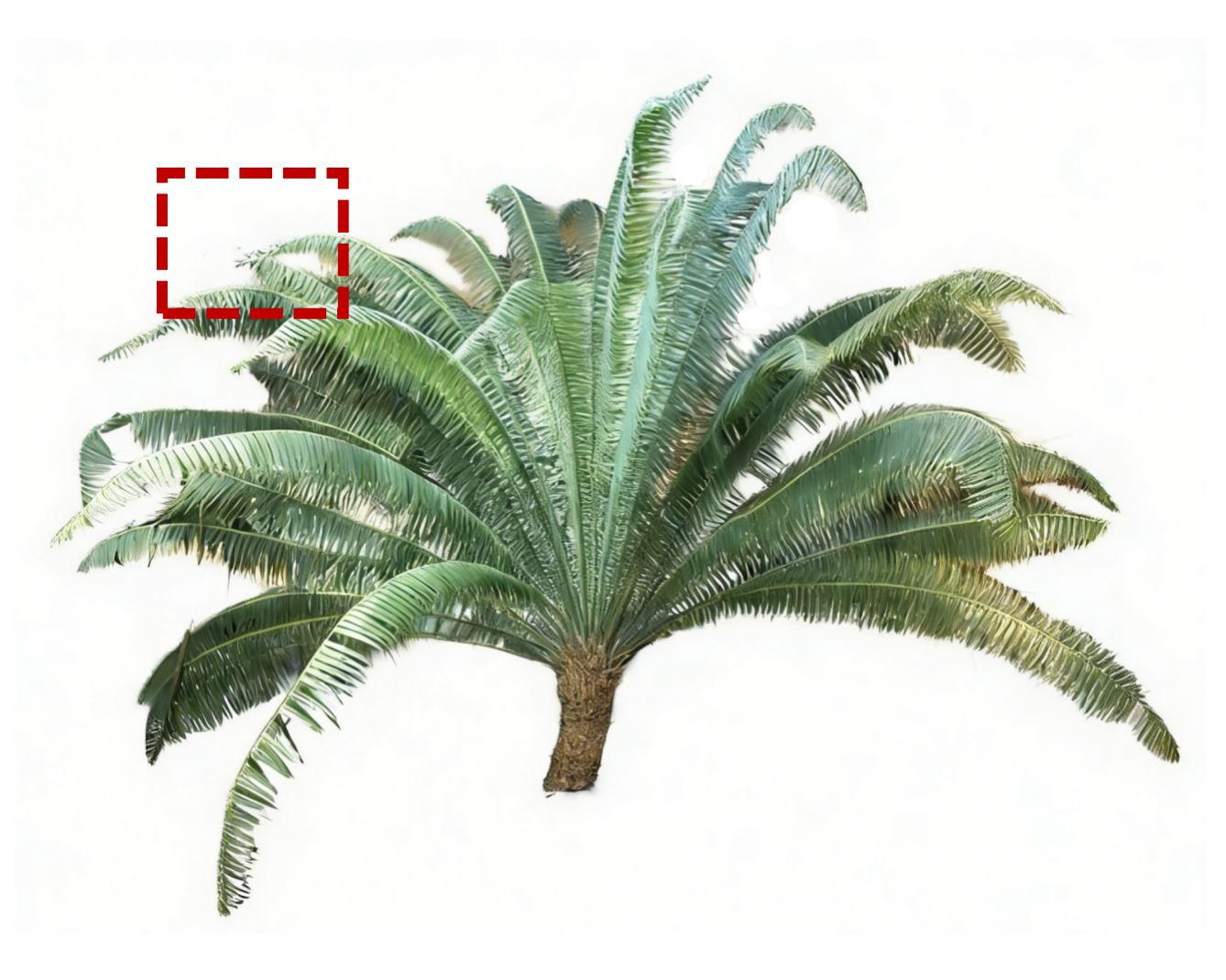}
        \caption{Our method}
        \label{fig:our_method}
    \end{subfigure}   
    \vspace{-0.4cm}
    \caption{Illustration robustness against erroneous masks.}
    \label{fig:noise 2D mask}
\end{figure}

Our method inherently addresses this concern via $\mathcal{L}_{\text{smth}}$ and RBF interpolation, and achieves more accurate results than COB-GS under the erroneous masks, showing in the Figure~\ref{fig:noise 2D mask}.
\section{Hyperparameter Experiment}
\label{Hyperparameter experiment}
\begin{figure}[H]
\centering
\vspace{-0.15cm}
\includegraphics[width=0.9\linewidth]{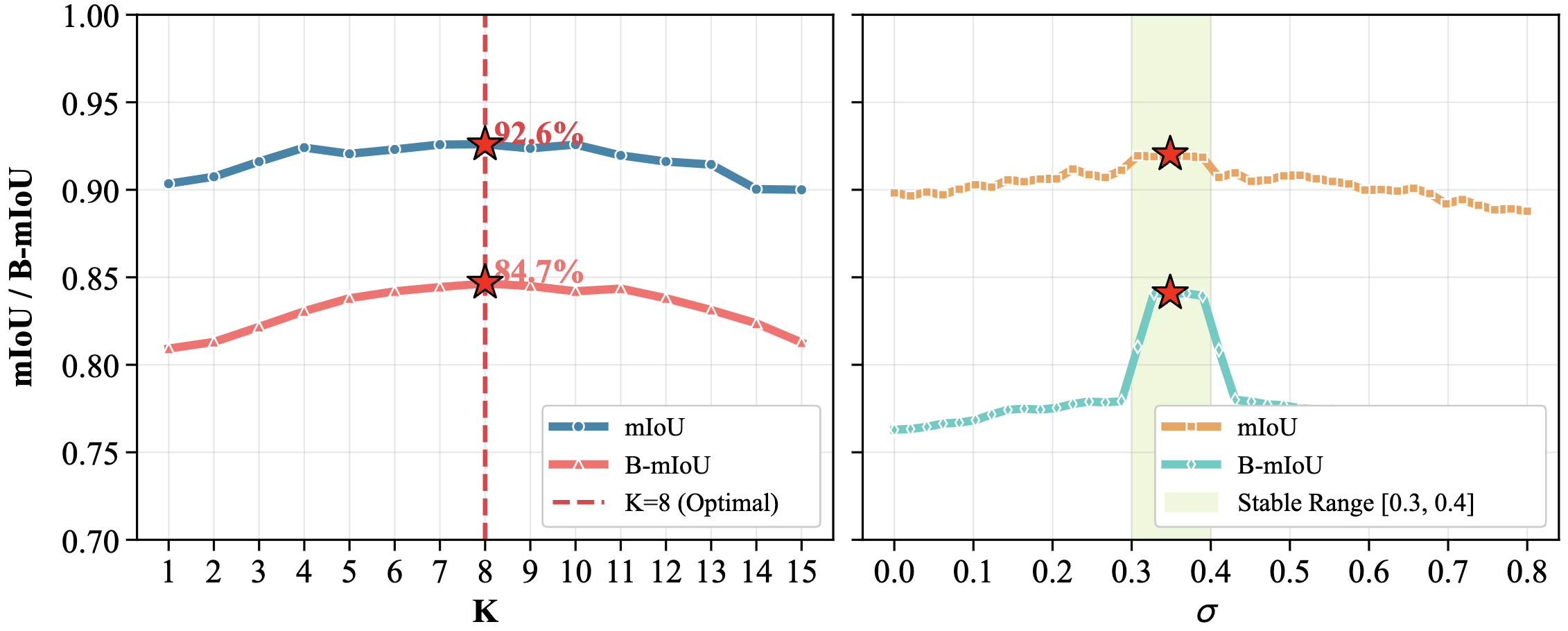} 
\caption{The impact of $\sigma$ and $K$ on mIoU and B-mIoU.}
\label{fig:parameter}
\end{figure}

As shown in Figure ~\ref{fig:parameter}, the parameter $\sigma$ remained stable within [0.3, 0.4], and $K=8$ was determined via grid search to effectively balance underfitting and noise.

\end{document}